%% file: rlc25.tex
\documentclass[10pt]{article} 

\usepackage[preprint]{rlj} 

\usepackage{amssymb}            
\usepackage{mathtools}          
\usepackage{mathrsfs}           
\usepackage{graphicx}           
\usepackage{subcaption}         
\usepackage[space]{grffile}     
\usepackage{url}                
\usepackage{lipsum}             

\usepackage{placeins}

\usepackage{amsmath}
\usepackage{amsfonts}
\usepackage{amssymb}
\usepackage{mathtools}
\usepackage{amsthm}

\usepackage{graphicx}
\usepackage{subcaption}        
\usepackage{wrapfig}           
\usepackage{float}             

\usepackage{booktabs}          
\usepackage{tabularx}          
\usepackage{multirow}          
\usepackage{makecell}          

\usepackage{algorithm}
\usepackage{algorithmic}

\usepackage{enumitem}          
\usepackage{soul}              
\usepackage{lipsum}            
\usepackage{pifont}            

\usepackage{url}               
\usepackage{microtype}         
\usepackage{nicefrac}          
\usepackage{xcolor}            
\usepackage[font=small]{caption}

\title{Sparse-Reg: Improving Sample Complexity in Offline Reinforcement Learning using Sparsity}

\setrunningtitle{Sparse-Reg: Improving Sample Complexity in Offline Reinforcement Learning using Sparsity}

\author{Samin Yeasar Arnob\textsuperscript{1,2}, Scott Fujimoto\textsuperscript{3}, Doina Precup\textsuperscript{1,2}}

\emails{samin.arnob@mail.mcgill.ca}

\affiliations{
$^{1}$\textbf{Department of Computing Science, McGill University, Canada}\\
$^{2}$\textbf{Mila Quebec AI Institute}\\
$^{3}$\textbf{Meta FAIR}
\par 
}

\contribution{
     This paper presents sparsity as a regularization technique to improve performance on small datasets in offline reinforcement learning within the continuous control domain, demonstrating that sparse regularization is a more reliable approach compared to other parameter regularization methods.
    }
    {
    Traditional sparsity literature primarily seeks to create the thinnest possible network to achieve faster inference and reduce computational overhead \citep{lottery_ticket, grasp, lee2018snip, arnob2021single, Rigl, rlx2}.
    }


\summary{In this paper, we investigate the use of small datasets in the context of offline reinforcement learning (RL). While many common offline RL benchmarks employ datasets with over a million data points, many offline RL applications rely on considerably smaller datasets. We show that offline RL algorithms can overfit on small datasets, resulting in poor performance. To address this challenge, we introduce "Sparse-Reg": a regularization technique based on sparsity to mitigate overfitting in offline reinforcement learning, enabling effective learning in limited data settings and outperforming state-of-the-art baselines in continuous control.
}

\begin{document}
\maketitle  

\begin{abstract}
In this paper, we investigate the use of small datasets in the context of offline reinforcement learning (RL). While many common offline RL benchmarks employ datasets with over a million data points, many offline RL applications rely on considerably smaller datasets. We show that offline RL algorithms can overfit on small datasets, resulting in poor performance. To address this challenge, we introduce "Sparse-Reg": a regularization technique based on sparsity to mitigate overfitting in offline reinforcement learning, enabling effective learning in limited data settings and outperforming state-of-the-art baselines in continuous control. \textbf{Codebase}: \href{https://github.com/SaminYeasar/sparse\_reg}{\url{https://github.com/SaminYeasar/sparse_reg}}
\end{abstract}

\input{files/01_introduction}

\input{files/03_background}
\input{files/02_related_work}
\input{files/04_proposed_method}

\input{files/06_results}
\input{files/07_conclusion}

\appendix

\input{files/08_appendix}


\newpage
\bibliography{rlc25}
\bibliographystyle{rlj}

\beginSupplementaryMaterials

\input{files/09_supplimentary}

\end{document}

%% file: files/01_introduction.tex
\section{Introduction}

Reinforcement Learning (RL) is a powerful approach to solving complex problems, but its application can be challenging due to the need for active interaction with the environment. In many scenarios, collecting training data can be challenging or resource-intensive. Offline RL offers an alternative by leveraging pre-existing datasets for training, avoiding any real-time interaction. 


Unsurprisingly, the dataset plays a significant role in the performance of an offline agent~\citep{schweighofer2021understanding, change_data_offline_RL, lambert2022challenges}
While most offline RL benchmarks emphasize large datasets (one million samples or more)~\citep{D4RL, gulcehre2020rl}, many application scenarios depend on learning from limited data samples. As a result, developing offline RL algorithms that perform well with small datasets is crucial for advancing the field of RL.





This paper examines offline RL with small datasets in continuous control and the use of sparsity to address limited data. We find the following core insights:


\textbf{Offline RL algorithms fail with small datasets:} Prior work in offline RL exhibits overfitting~\citep{REM, workflow_offlineRL, offline_RL_sample_complexity}, i.e., their performance starts to deteriorate after a certain number of gradient updates and this issue becomes severe with limited datasets~(under 100k samples)\citep{offline_RL_sample_complexity} in continuous control. Nevertheless, previous studies have not yet shown improvements in performance within \emph{limited} data settings.



\textbf{Sparse training addresses limited data:} To mitigate the risk of algorithms overfitting due to limited training data, we employ regularization techniques in the parameter space of neural networks. Drawing from techniques developed for pruning neural networks \citep{lecun1990optimal, hassibi1993optimal, dong2017learning, han2015learning, lottery_ticket, lee2018snip}, we reduce the complexity of the parameter space by inducing sparsity in the neural network. Our approach focuses on finding a balance between utilizing available data and preventing the model from becoming overly specialized. By carefully regulating the parameter space, we guide the model to capture essential patterns and information while avoiding the pitfalls of overfitting. Consequently, this strategy leads to improved performance in offline RL tasks, even when dealing with small training datasets. Through experiments conducted on continuous control tasks in the MuJoCo environment \citep{mujoco} within the D4RL benchmark \citep{D4RL}, we demonstrate the effectiveness of sparsity in dramatically improving sample complexity across various algorithms, dataset sizes, and tasks. Furthermore, we analyze different parameter regularization methods such as $L1$, Dropout \citep{dropout}, Weight Decay \citep{weight_decay}, Spectral Normalization \citep{spectral_norm}, Layer Norm \citep{layer_norm}. Our findings suggest that sparsity is a more reliable technique for enabling learning on small datasets.

The significance of this contribution lies in its potential to make offline RL more practical and applicable in scenarios where collecting high-quality samples may be challenging or prohibitively expensive. By harnessing regularization techniques and targeted parameter training, we aim to advance the field of offline RL and offer a promising avenue for researchers and practitioners to enhance the robustness and efficacy of offline RL algorithms.

%% file: files/03_background.tex
\section{Background}

We consider learning in a Markov decision process (MDP) described by the tuple ($S, A, P, R$). The MDP tuple consists of states $s \in S$, actions $a\in A$, transition dynamics $P(s'|s,a)$, and reward function $r=R(s,a)$. We use $s_t$, $a_t$ and $r_t=R(s_t,a_t)$ to denote the state, action and reward at timestep t, respectively. A trajectory comprises a sequence of states, actions and rewards $\tau=(s_0, a_0, r_0, s_1, a_1, r_1, ..., s_T,a_T,r_T)$. For continuous control task, we consider an infinite horizon, where $T=\infty$ and the goal in reinforcement learning is to learn a policy which maximizes the expected discounted return $\mathbb{E}[\sum_{t=t'}^T \gamma^t r_t]$ in an MDP.  
In offline RL~\citep{BCQ, REM}, instead of obtaining data through environment interactions, the agent learns from a fixed dataset of trajectory rollouts of arbitrary policies.

There is a range of pruning techniques for Deep Neural Networks that are proven to find a sparse network through an iterative update during training \citep{lecun1990optimal, HanPTD15, MolchanovTKAK16, Guo2016, liu2015sparse,lottery_ticket}. Recent studies on sparsity in online RL \citep{Rigl, rlx2} methods suggest leveraging gradient-based topology evolution criteria to identify sparse networks. However, these iterative methods require additional computation steps and additional training overhead. There are simpler alternatives that find the sparse network at a \emph{single-shot} \citep{lee2018snip, wang2020picking, synflow2020} during weight initialization and are proven to perform without loss of efficiency, with up to $95\%$ sparse network. These methods use gradient-based saliency criteria that rely on the Taylor expansion of the change in the loss to approximate the importance of the parameters. Thus considering the computation efficiency we use a single-shot pruning method to find important parameters at initialization.



%% file: files/02_related_work.tex

\section{Related Work}

Traditionally, the challenge of Offline RL focuses on mitigating extrapolation error~\citep{BCQ}, error introduced from uncertain samples. Offline RL algorithms address this concern by constraining the policy~\citep{wu2019behavior, AWAC, fisher_RBC, TD3_BC, fujimoto2023sale}, or penalizing or parametrizing the value function to avoid uncertain actions~\citep{CQL, garg2022extreme, IQL, hansen2023idql}. 

When considering the dataset, Offline RL research often focuses on enhancing the quality of data samples \citep{data_centric_offline_RL,  change_data_offline_RL}. However, this is impractical and costly for real-world applications. One major challenge with Offline RL is that the algorithms perform poorly when the training dataset is small~\citep{offline_RL_sample_complexity}.
Previous research, such as \citep{REM} and \citep{workflow_offlineRL}, has explored and confirmed the occurrence of overfitting in offline RL but not in the context of data-constrained settings. In this paper, we show that regularization techniques can be useful in tackling such issues. We propose to use sparsity in the neural network as a regularization method to solve this issue and overcome the overfitting problem.

Our approach to incorporating sparsity differs from prior works in the sparsity literature in both supervised learning \citep{lottery_ticket, synaptic_pruning, gale2019state, blalock2020state, yang2019deephoyer, lee2018snip, ohib2022explicit} and RL \citep{arnob2021single, arnob2025efficient, rlx2}. Traditional sparsity literature primarily seeks to create the thinnest possible network to achieve faster inference and reduce computational overhead. In contrast, this paper's utilization of sparsity is centred on a distinct purpose: \emph{the regularization of neural networks to decrease model complexity, particularly when dealing with limited training data samples.}



%% file: files/04_proposed_method.tex
\section{Sparse-Reg: Sparse Regularization for Offline RL}


We propose using sparse regularization to regularize model complexity when learning from small datasets. We use a gradient-based method proposed in \cite{lee2018snip} that identifies important parameters by using a \emph{connection sensitivity} measure $S$, defined as the influence of weight on the loss function. They formalize this idea in terms of removing individual weights $\theta_q$, and the effect it has on the loss deriving the following sensitivity score: $S(\theta_q) = \lim_{ \epsilon \to 0} \left| \frac{\mathcal{L}(\theta) -\mathcal{L} (\theta + \epsilon \delta_q)}{\epsilon} \right| = \left| \theta_q \frac{\partial \mathcal{L}}{\partial \theta_q} \right|$, where $\delta_q$ is a vector whose $q_{th}$ element equals $\theta_q$ and all other elements are $0$. Thus, to select the important parameter, we use the following objective:
8
\begin{equation} \label{eq:snip}
\scalebox{0.8}{$\begin{aligned}
 m = & T_k(S(\theta_q; D)), \\
   = & T_k \left( \lim_{ \epsilon \to 0} \left| \frac{\mathcal{L}(\theta; D) -\mathcal{L} (\theta + \epsilon \delta_q; D)}{\epsilon} \right| \right), \\
   = & T_k \left( \left| \theta_q \frac{\partial \mathcal{L}(\theta; D)}{\partial \theta_q} \right| \right), 
\end{aligned}$}
\end{equation}

where $T_k$ is defined as the “top$-$k” operator, the function identifies the “top$-$k” parameter and sets the positional value as 1 and the rest as 0, allowing us to have a matrix $m$ which is used as a mask on the parameters~$\theta$. Formally, we consider a feed-forward network $f(x,\theta)$ with initial parameters $\theta \sim \Theta$. 

We apply Sparse-Reg by selecting a subset of parameters $f(x,\theta \odot m)$ by using a mask $m \in \{0,1\}^{|\theta|}$ derived through Equation \eqref{eq:snip}. The regularization is governed by sparsity, which refers to the proportion of parameters that are pruned and set to zero. By increasing the sparsity percentage, we decrease the number of trainable parameters, thereby reducing the model's complexity.

Offline RL algorithms employ several deep neural networks, each serving distinct objective functions. For instance, within an actor-critic framework, we have an actor network denoted as $\pi(\theta)$ and a critic network denoted as $Q(\phi)$, each equipped with its unique objective function. Thus for each neural network, we use the scoring function in Equation (\ref{eq:snip}) to find masks ($m^{\theta},m^{\phi}$) for each network. We multiply the found mask to get a regularized model $\pi(\theta*m^{\theta})$, $Q(\phi*m^{\phi})$. Several RL algorithms incorporate either a target-actor or target-critic network, and occasionally both. These target networks are initially set with identical weights but receive delayed updates. Consequently, we utilize the mask intended for the current actor-critic networks to apply to the target networks as well.

We take a batch of data from the offline training dataset $D$, calculate the gradient through forward and backward propagation, and then use Equation \eqref{eq:snip} to generate masks for the neural networks. Periodic update of sparsity is found to be more effective for RL \citep{arnob2025efficient}. To achieve the best-regularized model using sparse regularization, we aim to apply sparsity repeatedly while updating the associated parameters.  
In every 5k gradient steps, we periodically update the masks only for 200k training steps. Through an ablation study, we demonstrate the impact of applying periodic sparse regularization in Section \ref{sec:ablation}.

A code snippet to integrate the method into the offline RL algorithms is provided in Figure \ref{fig:code_snippet}. 

%% file: files/06_results.tex
\begin{figure*}[!htb]
\vspace{-10pt}
\centering
\scalebox{0.8}{%
  \begin{subfigure}[b]{0.3\textwidth}
    \includegraphics[width=\linewidth]{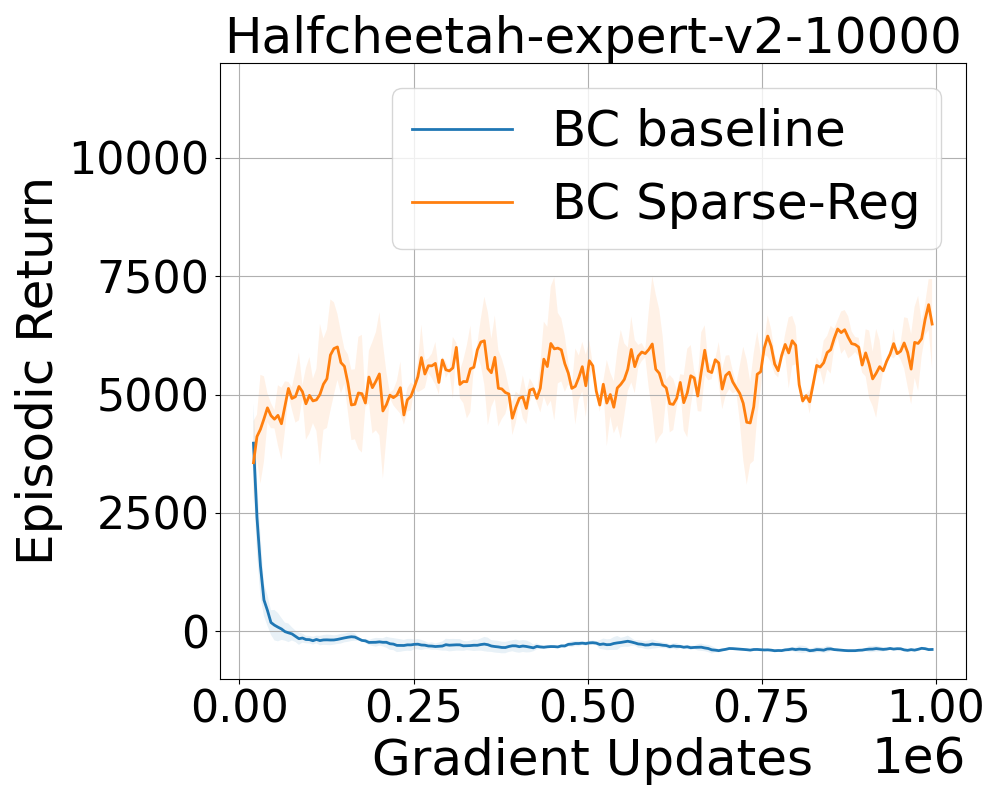}
    \caption{}
  \end{subfigure}
  \begin{subfigure}[b]{0.3\textwidth}
    \includegraphics[width=\linewidth]{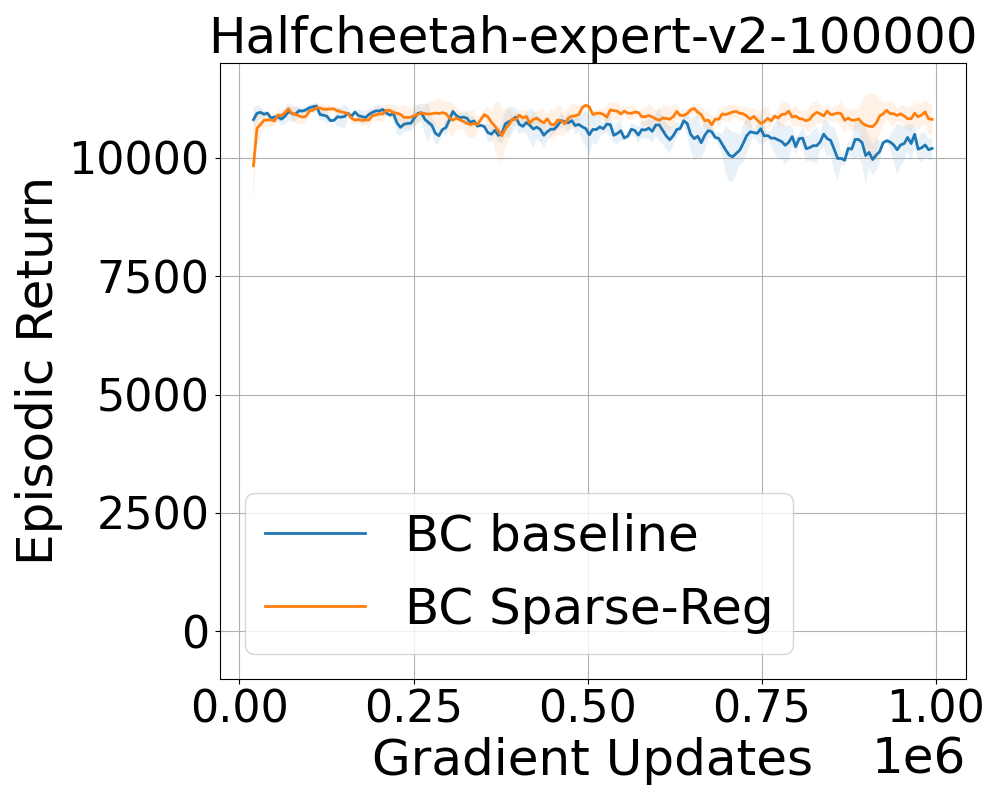}
    \caption{}
  \end{subfigure}
  \begin{subfigure}[b]{0.3\textwidth}
    \includegraphics[width=\linewidth]{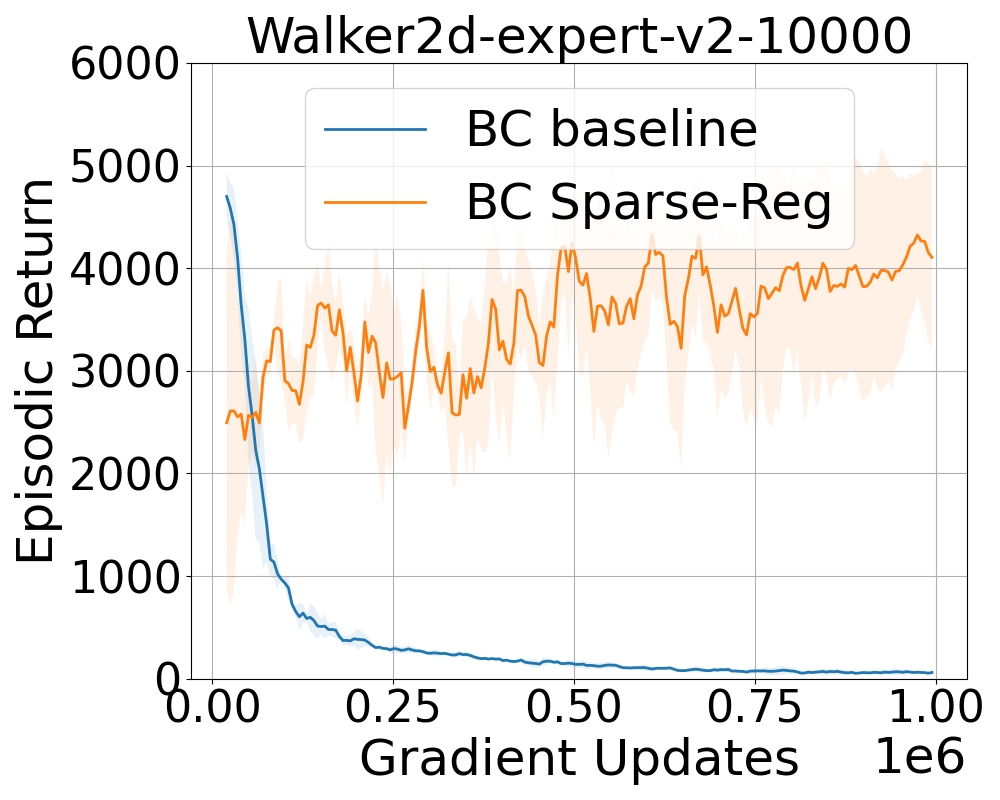}
    \caption{}
  \end{subfigure}
  \begin{subfigure}[b]{0.3\textwidth}
    \includegraphics[width=\linewidth]{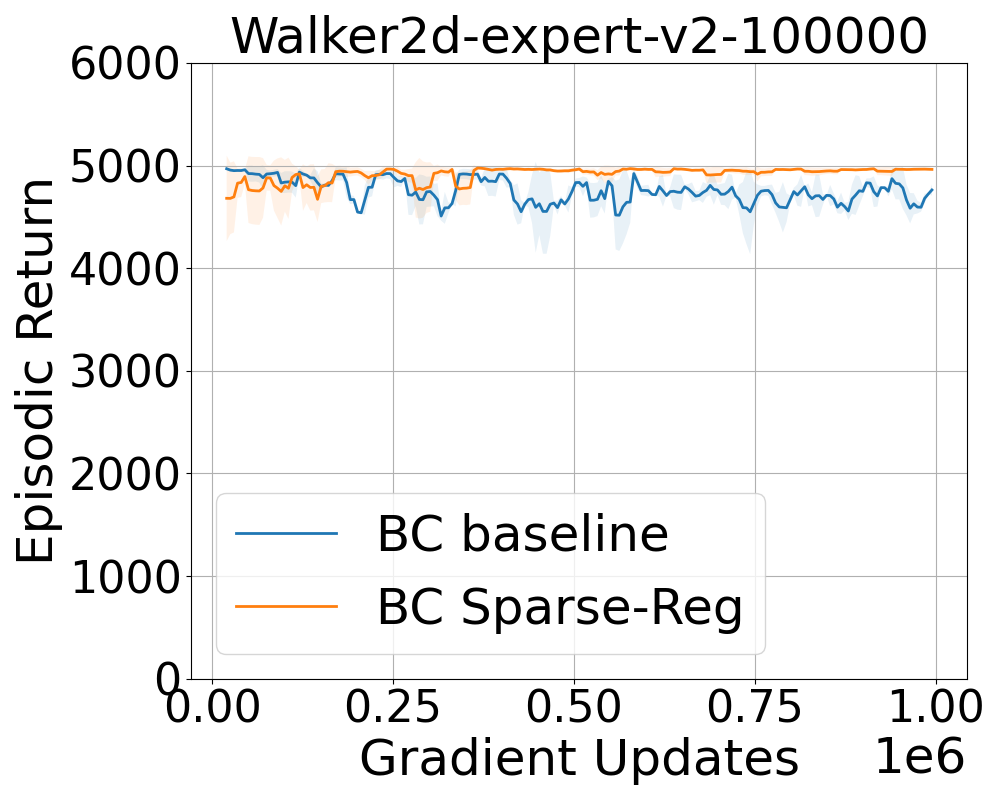}
    \caption{}
  \end{subfigure}
}
\caption{Learning curve of BC-baseline (\textcolor{cyan}{blue}) and BC-sparse (\textcolor{orange}{orange}) for \texttt{HalfCheetah-v2} with ($a$) $10k$ and ($b$) $100k$ training samples; \texttt{Walker2d-v2} with ($c$) $10k$ and ($d$) $100k$ training samples over 1 million time steps. Performance mean and standard deviation of the episodic returns are estimated every $5k$ steps over 10 evaluations across 5 seeds.}
\vspace{5pt}
\label{fig:BC_baseline_learning_curve}
\end{figure*}

\section{Experiments}

In this section, we provide empirical evidence supporting the effectiveness of network sparsity regularization in mitigating overfitting and enhancing the performance of various offline RL algorithms. Additionally, we conduct a comparative analysis with alternative regularization techniques, demonstrating the superior reliability of our approach.

We conduct our experiments using datasets from D4RL \citep{D4RL}. We present experimental results for \emph{Expert} in the main text, while additional results conducted on \emph{Medium}, \emph{Medium-Replay}, and \emph{Expert-Replay} are provided in the Appendix.  For sample complexity analysis, we compare the performance under varying sample sizes. Contrary to offline RL algorithm benchmarks \citep{D4RL, IQL, TD3_BC} with over 1 million training samples, our experiments are conducted in limited dataset settings, employing sample sizes of $5k$, $10k$, $50k$, and $100k$. We simultaneously enhance model complexity when increasing the training dataset to $50k$ and $100k$ samples. When dealing with smaller sample sizes like $5k$ and $10k$, we train using only $5\%$ of the parameters ($95\%$ sparse). However, for larger sample sizes such as $50k$ and $100k$, we choose to utilize $25\%$ of the parameters ($75\%$ sparse).  We justify our choice of this hyper-parameter in the ablation study in Section \ref{sec:ablation_sparsity_ratio}. In all experiments, we conducted training with 1 million gradient updates (if not specifically stated) and reported the mean and standard deviation results obtained across 5 seeds. In this paper, we also use quantile bar plots for more compact visualization, illustrating the model's final evaluation results (mean of 10 episodes) across 5 seeds. 

\subsection{Behavior Cloning with Sparse-Reg}


We initiate our experiment using the \emph{Behavior Cloning (BC)} algorithm due to its simplicity, featuring a single policy network trained on an offline dataset using the mean-square-error (MSE) loss function. Figure \ref{fig:BC_baseline_learning_curve} illustrates the performance comparison of BC baseline and sparse regularized version (BC-sparse) on \texttt{HalfCheetah-v2} and \texttt{Walker2d-v2} tasks. We assess the episodic return every $5k$ step throughout 1 million gradient steps. We show this learning curve across two experimental setups employing $10k$ and $100k$ training samples. The disparity in performance between the BC-baseline and the BC-sparse method is clear from Figure \ref{fig:BC_baseline_learning_curve}, with BC-sparse demonstrating superior performance. Additional experimental results with varying training samples are presented in Appendix Figure \ref{fig:appendix:BC_baseline_learning_curve}.



\input{files/highlight_table_new_results}


\subsection{Performance Evaluation of Offline RL Baselines}

We perform a performance evaluation comparing various Offline RL baselines: AWAC \citep{AWAC}, TD3+BC \citep{TD3_BC} and IQL \citep{IQL} with their sparse counterparts, as displayed in Table \ref{tab:offlineRL_baseline}. We utilize the \emph{Expert} dataset for this comparison. Here, we present the final evaluation results of offline algorithms alongside the respective performance of sparse regularized models following 1 million training steps. We calculate the mean and standard deviation of the final episodic return across 10 episodes for 5 seeds. This comparison highlights that our method significantly improves performance across the entire range of training samples. Additionally, the results for the \emph{Medium} data are provided in the Appendix in Table \ref{tab:offlineRL_baseline_medium}.  

\begin{figure}
\centering
\scalebox{0.7}{
\includegraphics[width=0.33\linewidth]{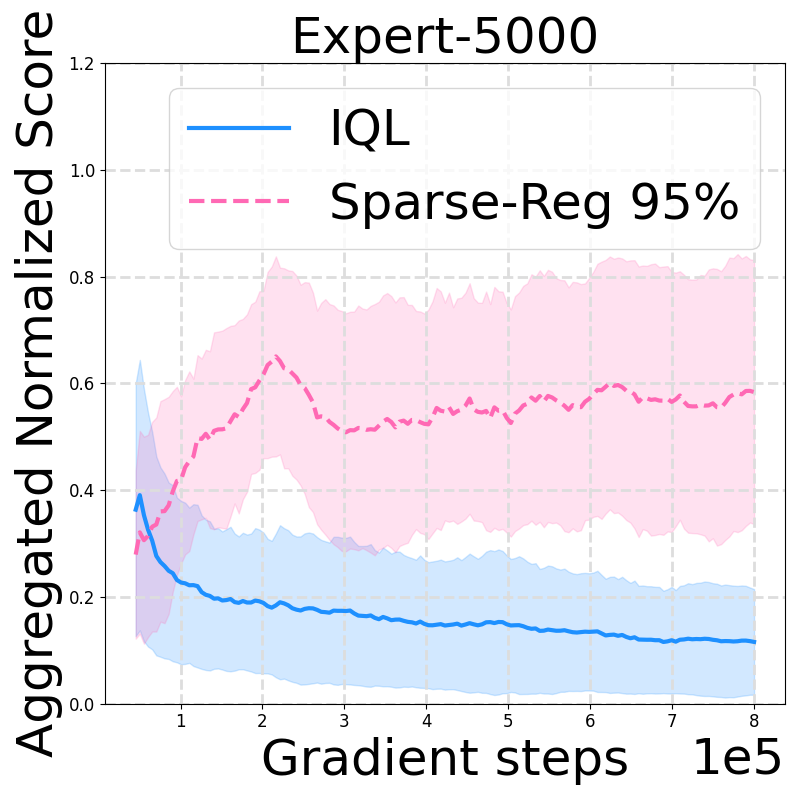}
\includegraphics[width=0.33\linewidth]{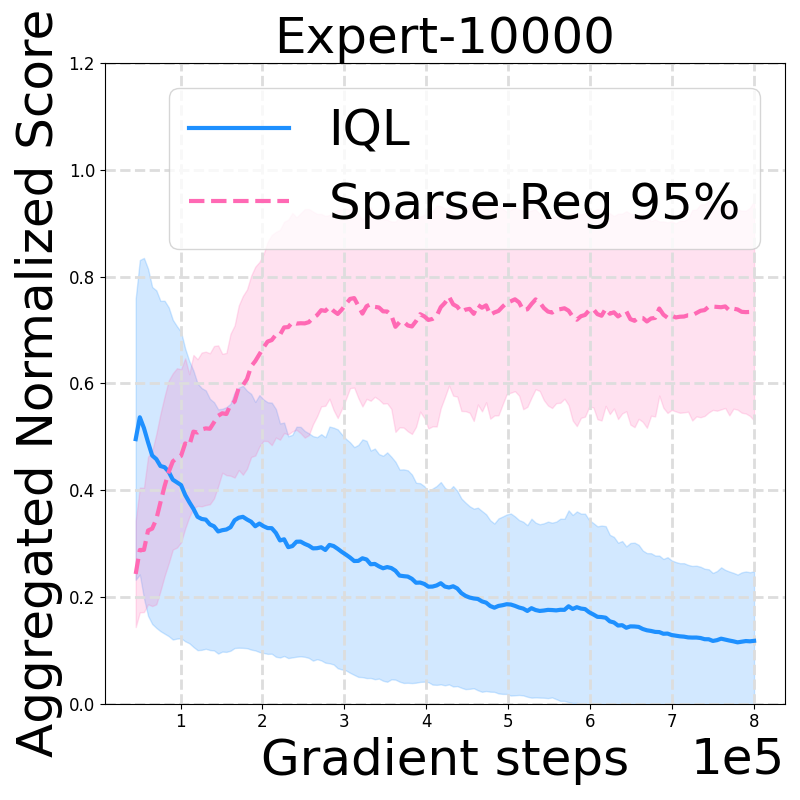}
\includegraphics[width=0.33\linewidth]{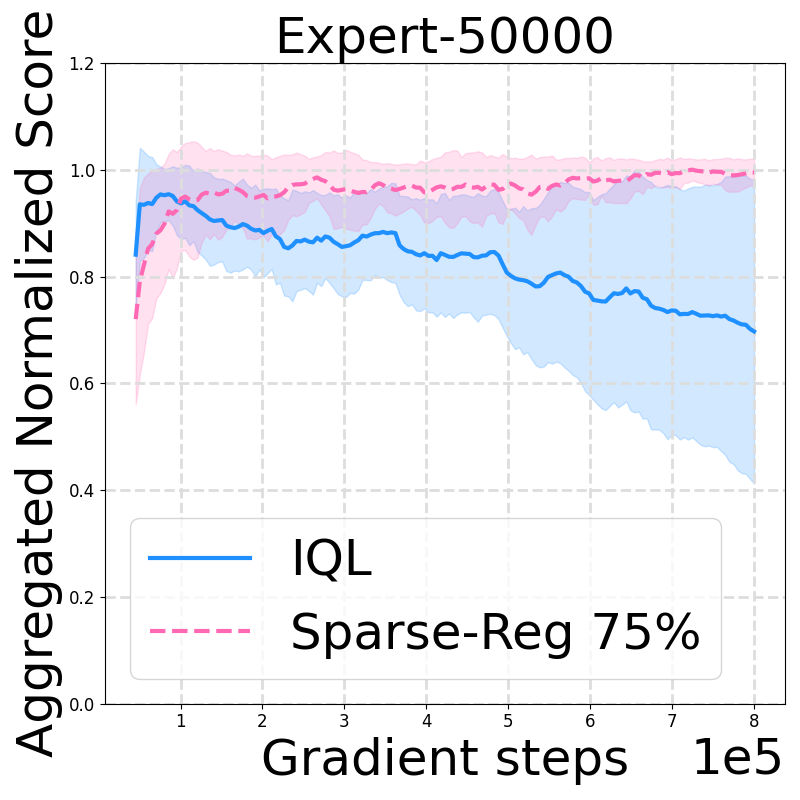}
\includegraphics[width=0.33\linewidth]{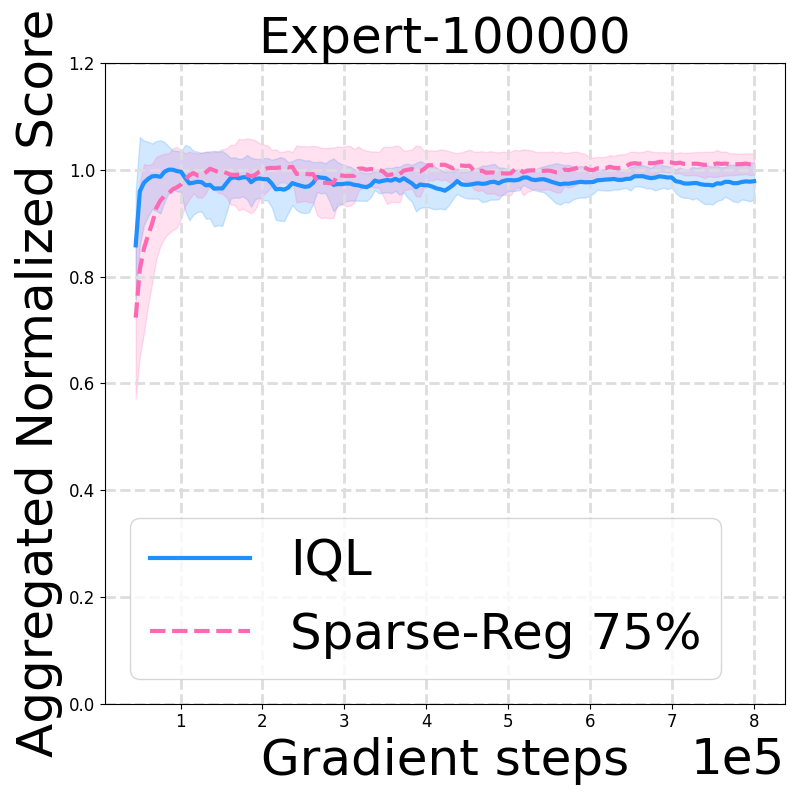}
 } 
  \vspace{-6pt}
  \caption{Aggregated normalized performance of IQL and IQL-Sparse-Reg across the \texttt{HalfCheetah}, \texttt{Hopper}, and \texttt{Walker2d} environments with varying training sample sizes (5k, 10k, 50k, and 100k) on D4RL-\emph{Expert} dataset. For the 50k and 100k training samples, we increase the trainable parameters by reducing sparsity from 95\% to 75\%}
\label{fig:agg_IQL_expert_mujoco}
\end{figure}

To enhance clarity in visualization, Figure \ref{fig:agg_IQL_expert_mujoco} illustrates instances where we improve performance through sparse regularization.  We first normalize the episodic return using standard proposed metric from \cite{D4RL}: $\text{Normalized score} = \big( \frac{\text{score - random score}}{\text{expert score -  random score}} \big)$, which scales the episodic return for to 1. After obtaining the normalized return for each environment, method and seed, we aggregate them over \texttt{HalfCheetah}, \texttt{Hopper}, and \texttt{Walker2d} tasks. Figure \ref{fig:agg_IQL_expert_mujoco} the aggregated normalized performance over varying training sample sizes: 5k, 10k, 50k, and 100k on \emph{Expert} data sample. Additionally, the results for the \emph{Medium} data are provided in the Appendix Figure \ref{fig:agg_IQL_medium_mujoco}.  



\subsection{Comparing Performance with Various Regularization Techniques}
\begin{figure}
\vspace{-10pt}
\centering
\scalebox{0.7}{
    \includegraphics[width=0.33\textwidth]{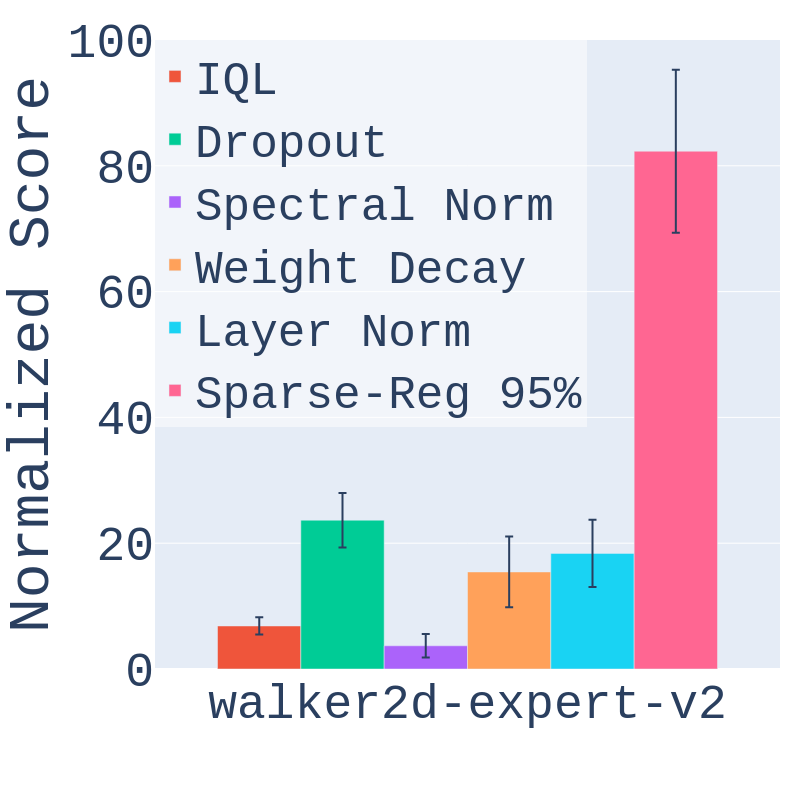}
    \includegraphics[width=0.33\textwidth]{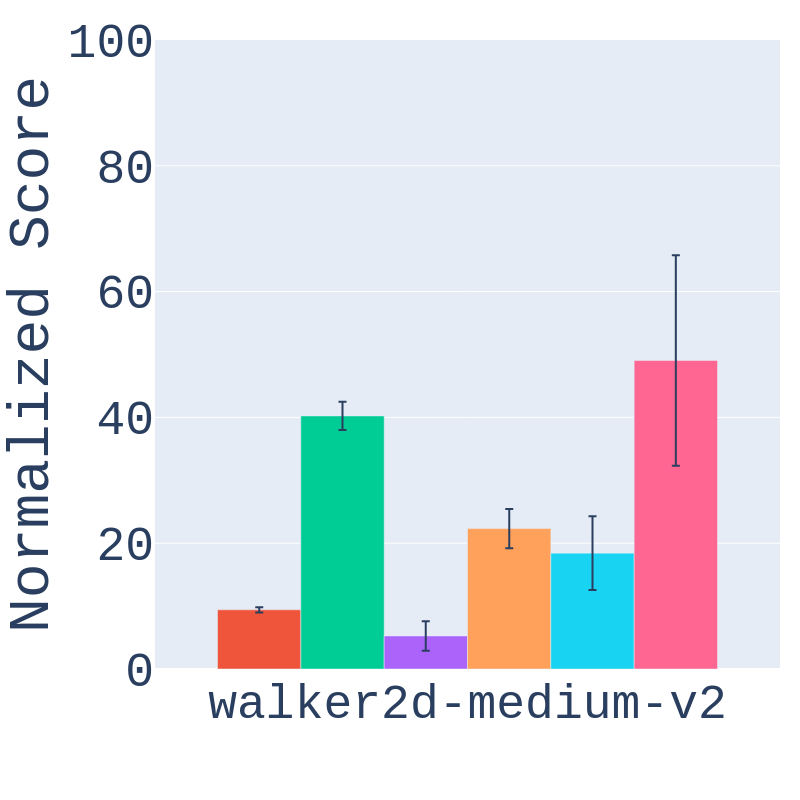}
  }
\!
\scalebox{0.7}{
    \includegraphics[width=0.33\textwidth]{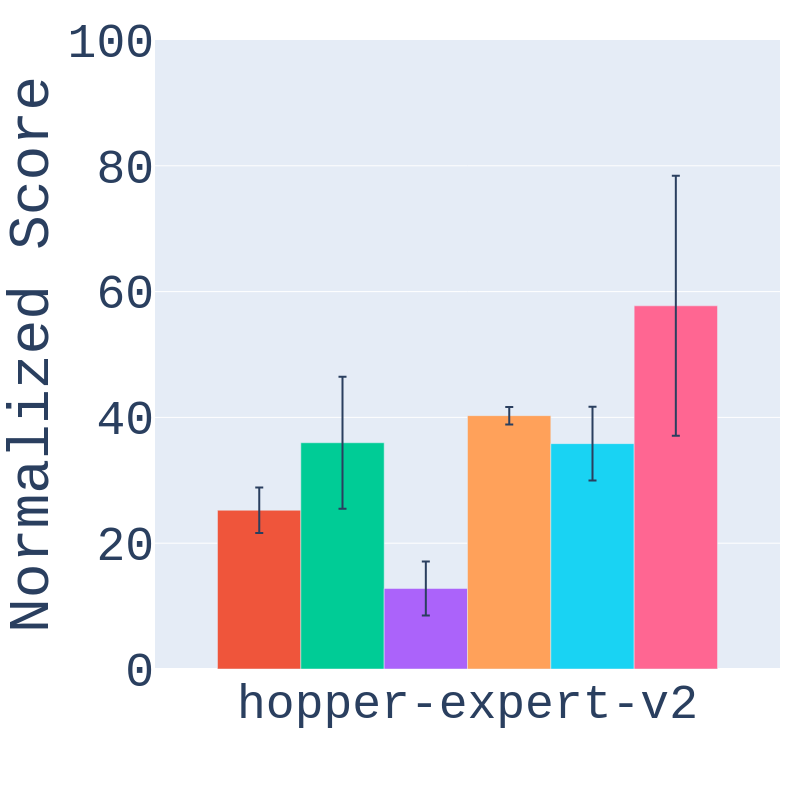}
    \includegraphics[width=0.33\textwidth]{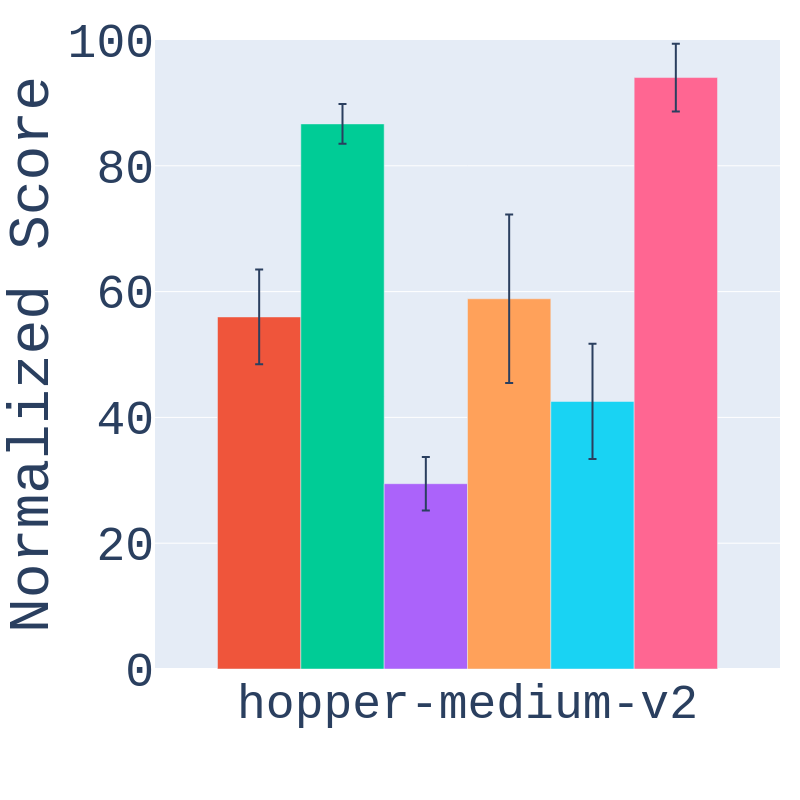}
  }
  \caption{Comparison of normalized performance for various regularizers applied to IQL algorithm, trained with 5k Samples.}
  \vspace{-10pt}
  \label{fig:IQL_other_reg_compare}
\end{figure} 
We hypothesize that using sparse regularization enables a more controlled and targeted regularization of neural network parameters. We evaluate a range of alternate regularization techniques designed to address overfitting in neural networks. These regularizers include dropout \cite{dropout} weight decay \cite{weight_decay}, spectral normalization \cite{spectral_norm}, layer norm \cite{layer_norm}. We compare the normalized scores for Walker2d on the Expert and Medium datasets, each with 5k training samples. We apply these regularization methods to IQL and present the results in Figure \ref{fig:IQL_other_reg_compare}.

We conducted four experiments on the \texttt{Walker2d} and \texttt{Hopper} environments with 5k training samples and found that the sparse method consistently outperforms the other methods. 

\begin{figure*}[t]
\vspace{-10pt}
\centering
\scalebox{0.8}{%
  \begin{subfigure}[b]{0.32\textwidth}
    \includegraphics[width=\linewidth]{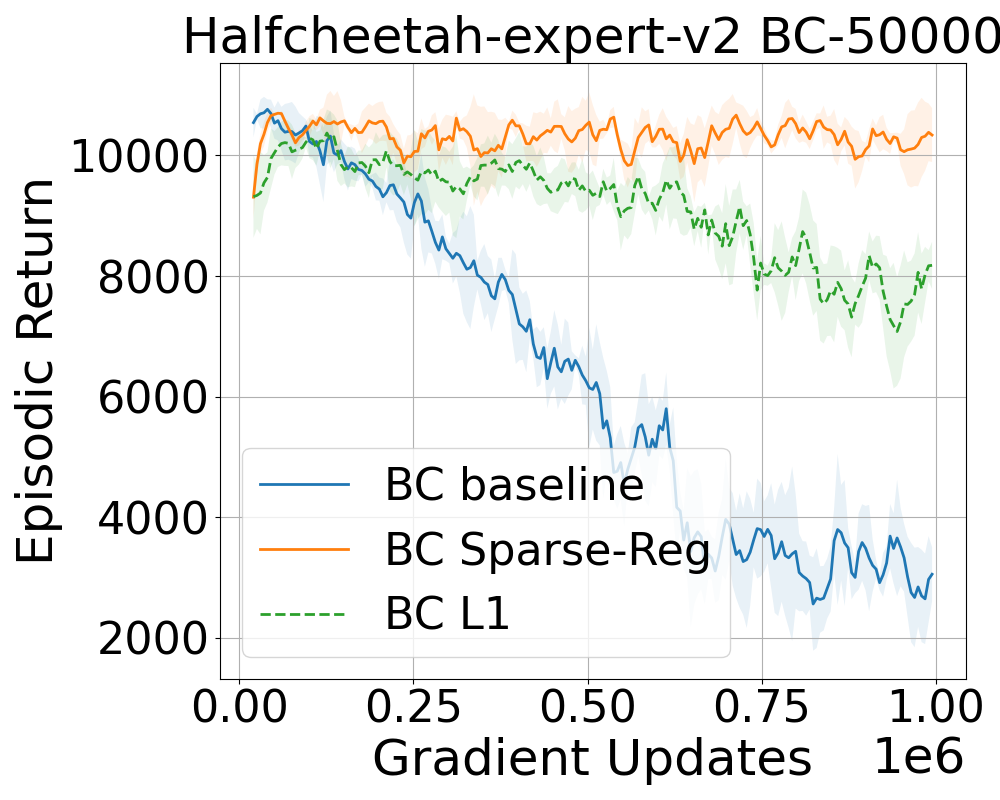}
    \caption{Episodic Return}
  \end{subfigure}
  \hspace{0.01\textwidth}
  \begin{subfigure}[b]{0.32\textwidth}
    \includegraphics[width=\linewidth]{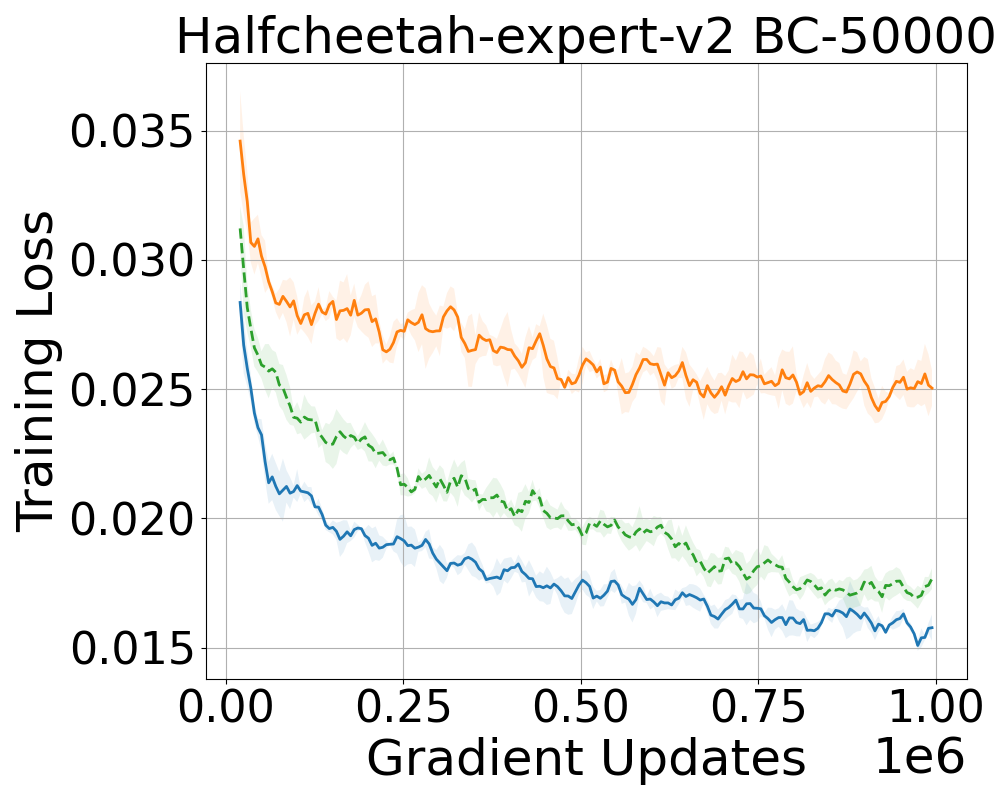}
    \caption{Training Loss}
  \end{subfigure}
  \hspace{0.01\textwidth}
  \begin{subfigure}[b]{0.32\textwidth}
    \includegraphics[width=\linewidth]{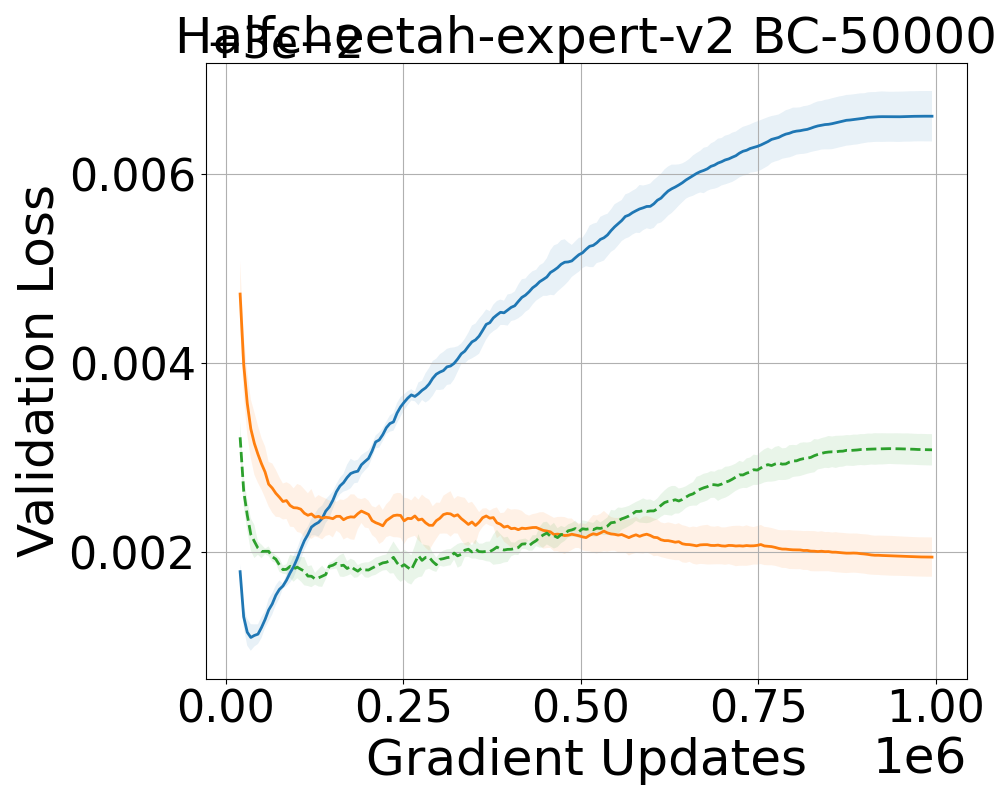}
    \caption{Validation Loss}
  \end{subfigure}
}
\caption{To investigate overfitting, we compare the performance of BC-baseline (\textcolor{cyan}{blue}) with sparse (\textcolor{orange}{orange}) and L1 (\textcolor{green!70!black}{green}) regularized BC. Along with the ($a$) episodic-return, we compare the ($b$) MSE loss of behaviour-action prediction on the \textbf{training dataset} and ($c$) the \textbf{validation dataset} over 1 million training steps. We train the algorithm with 50k training samples.}
\vspace{-10pt}
\label{fig:overfit_loss}
\end{figure*}

\subsection{Empirical Evidence of Reduced Overfitting}
    
    

We provide empirical evidence demonstrating that, while a naive offline RL method tends to fall prey to overfitting, targeted parameter training using sparsity effectively acts as a regularization technique, leading to improved performance. 

To investigate overfitting in offline RL algorithms, we perform a controlled experiment by training a policy network solely using behaviour cloning in a supervised learning framework. We propose to use a separate validation dataset. We held-out $200k$ $\{s_{V},a_{V},r_{V},s^{'}_{V}\}$ tuples from the D4RL dataset \citep{D4RL} during training. We perform evaluation over the validation dataset, which provides an unbiased and generalized estimation of the learning agent.


\begin{wrapfigure}{r}{0.6\linewidth}
\vspace{-10pt}
\centering
\scalebox{0.9}{%
  \begin{subfigure}[b]{0.3\textwidth}
    \includegraphics[width=\linewidth]{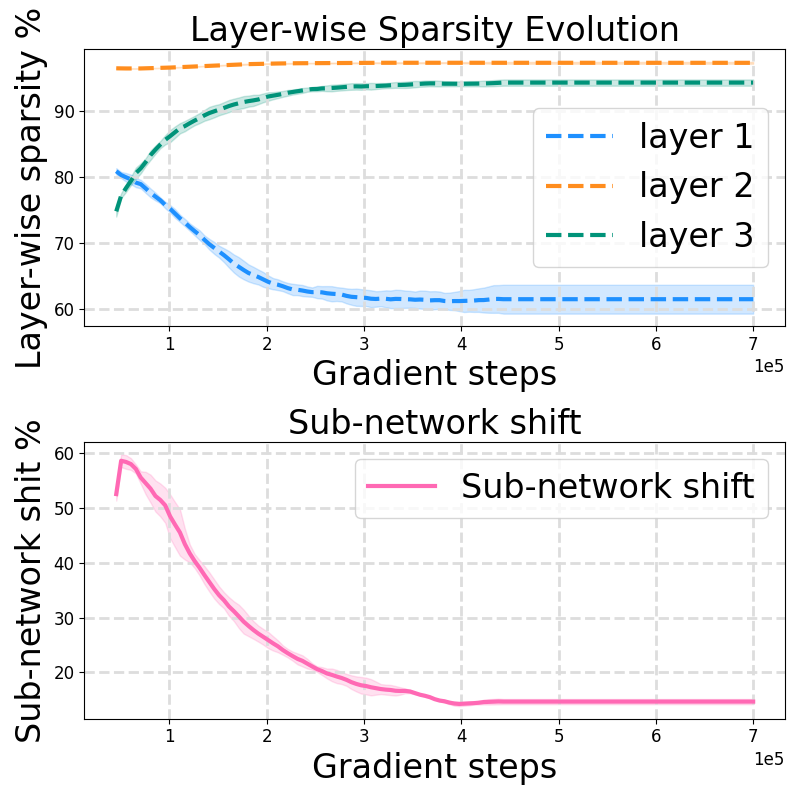}
    \caption{}
  \end{subfigure}
  \hspace{10pt}
  \begin{subfigure}[b]{0.3\textwidth}
    \includegraphics[width=\linewidth]{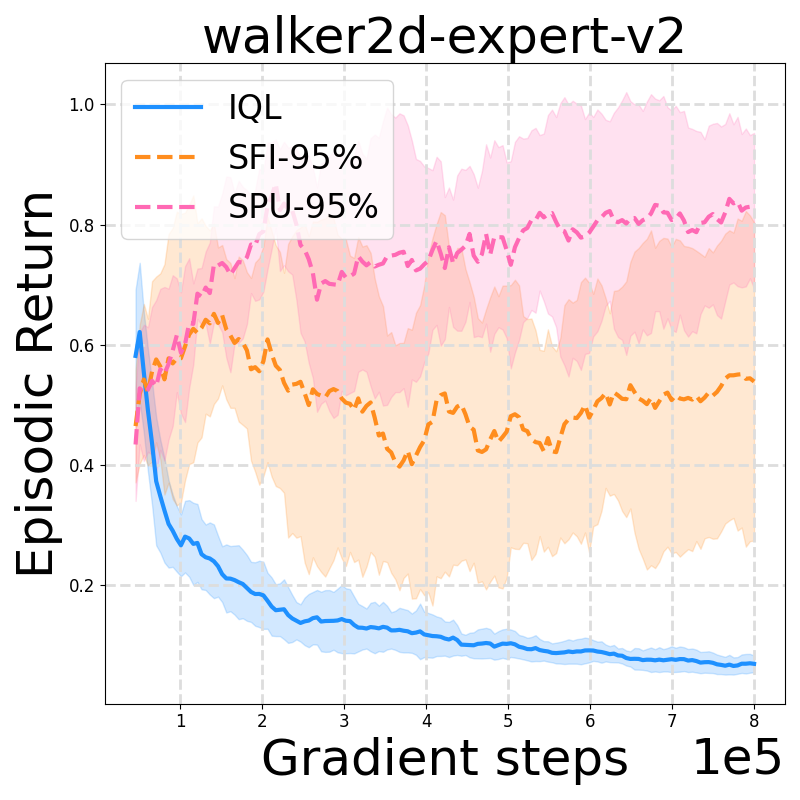}
    \caption{}
  \end{subfigure}
}
\caption{(a) Impact of applying network sparsity periodically during training. \textbf{Top}: Layer-wise sparsity evaluation. \textbf{Bottom}: Percentage change in sub-network parameters. (b) Learning curves for IQL and its sparse regularization variants: (i) sparse fixed at initialization (SFI) and (ii) sparse with periodic updates (SPU), where sparse regularization is applied every 5k gradient updates for the first 200k gradient steps.}
\vspace{-10pt}
\label{fig:SFI_SPU}
\end{wrapfigure}

As an evaluation criterion, we use the Mean-Square-Error (MSE) loss between behaviour-action $a$ and policy-action $\pi_\theta(s)$ to measure the actor's deviation from the distribution of the offline dataset. We compute $MSE(a_T,\pi_\theta(s))$ to assess the training loss, where $a_T$ refers to the training dataset. For validation loss, we calculate $MSE(a_V, \pi_\theta(s))$, with $a_V$ representing the validation dataset. In our experiments, both training and validation datasets are collected from the same behaviour policy distribution and thus an unbiased trained policy is expected to reduce the validation loss as well. Given the limited size of our training samples, we confine our experiments to a single behaviour policy. 
Using a validation score across various distributions could potentially lead to erroneous conclusions about the policy performance.

For this experiment, we compare the performance of the BC baseline with sparse and L1 \citep{L11996regression} regularized BC. In Figure \ref{fig:overfit_loss}(b), it is evident that BC exhibits a considerably greater reduction in policy training loss, followed by BC-L1 compared to BC-sparse. However, when examining the validation performance in Figure \ref{fig:overfit_loss}(c), BC experiences an overshoot of the loss. This represents a textbook case of overfitting \citep{bishop2006pattern}, where the declining training loss can misleadingly suggest a performance improvement.  While L1 regularization partially mitigates the issue of overshooting, we achieve the minimum validation error using sparse regularization. The correlation between minimizing validation loss and actual performance is evident from Figure \ref{fig:overfit_loss}($a$).

Based on the validation action prediction loss, we can conclude that sparse networks effectively reduce overfitting on the training data, enhancing performance generalization, as illustrated in Figure \ref{fig:overfit_loss}. 


\subsection{Ablation Study}
\label{sec:ablation}

\paragraph{Effects of Periodic Sparsity on Model Performance:}



To obtain the best-regularized model using Sparse-Reg, we aim to iteratively apply sparsity while updating the associated parameters. In our controlled experiments, we employ a scoring function (Equation \ref{eq:snip}) to sample subnetworks multiple times throughout training. Each time we sample a subnetwork, the scoring function is evaluated across the entire parameter space. This results in a dynamic sparsity distribution, with shifts occurring across different layers as the training progresses. Initially, we observe that the first layer is sparser, but its sparsity decreases as training advances. This shift in sparsity is expected and ultimately stabilizes. We conduct an experiment to understand the influence of periodic sparse update. For \emph{sparse fixed at initialization} (SFI), we apply sparsity only at the parameter initialization stage
In contrast, for \emph{sparsity with periodic updates} (SPU), we periodically update the subnetwork by recalculating the importance of connections during the first 200k steps of training, allowing us to identify and refine a better subnetwork over time.

In Figure \ref{fig:SFI_SPU}$(a)$ we illustrate the evolution of layer-wise sparsity in the IQL actor network when sparsity is periodically applied throughout training. For an actor-network with two hidden layers, the initial layer becomes less sparse as the scoring function identifies its parameters as more important, leading to increased sparsity in the outer layer. This adjustment in parameter distribution across layers appears to affect performance as well.

Heuristically, we find that in our experiments, the sparsity ratio stabilizes after 200k gradient steps, although there is a certain subnetwork shift throughout the training (see Figure \ref{fig:SFI_SPU}$(a)$, bottom). Hence, we stop the periodic updates after 200k steps. In Figure \ref{fig:SFI_SPU}$(b)$, we observe an improvement with periodic update compared to fixed at initialization, which we apply for all our Sparse-Reg experiments.

\paragraph{Impact of Varying Sparsity:}
\label{sec:ablation_sparsity_ratio}


\begin{wrapfigure}{r}{0.6\linewidth}
\vspace{-10pt}
\centering
\scalebox{0.85}{%
  \begin{subfigure}[b]{0.29\textwidth}
    \includegraphics[width=\linewidth]{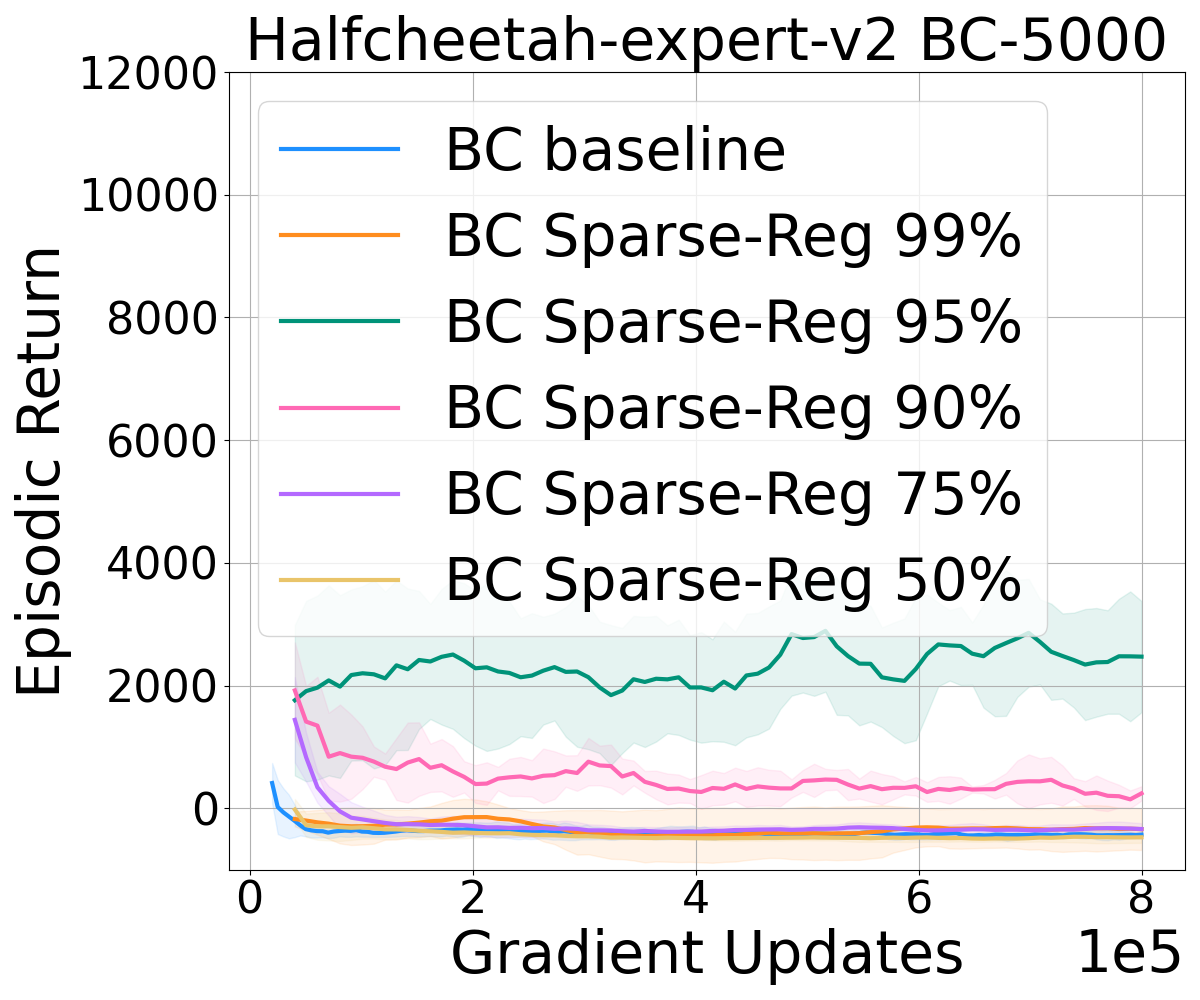}
    \caption{}
  \end{subfigure}
  \hspace{3pt}
  \begin{subfigure}[b]{0.29\textwidth}
    \includegraphics[width=\linewidth]{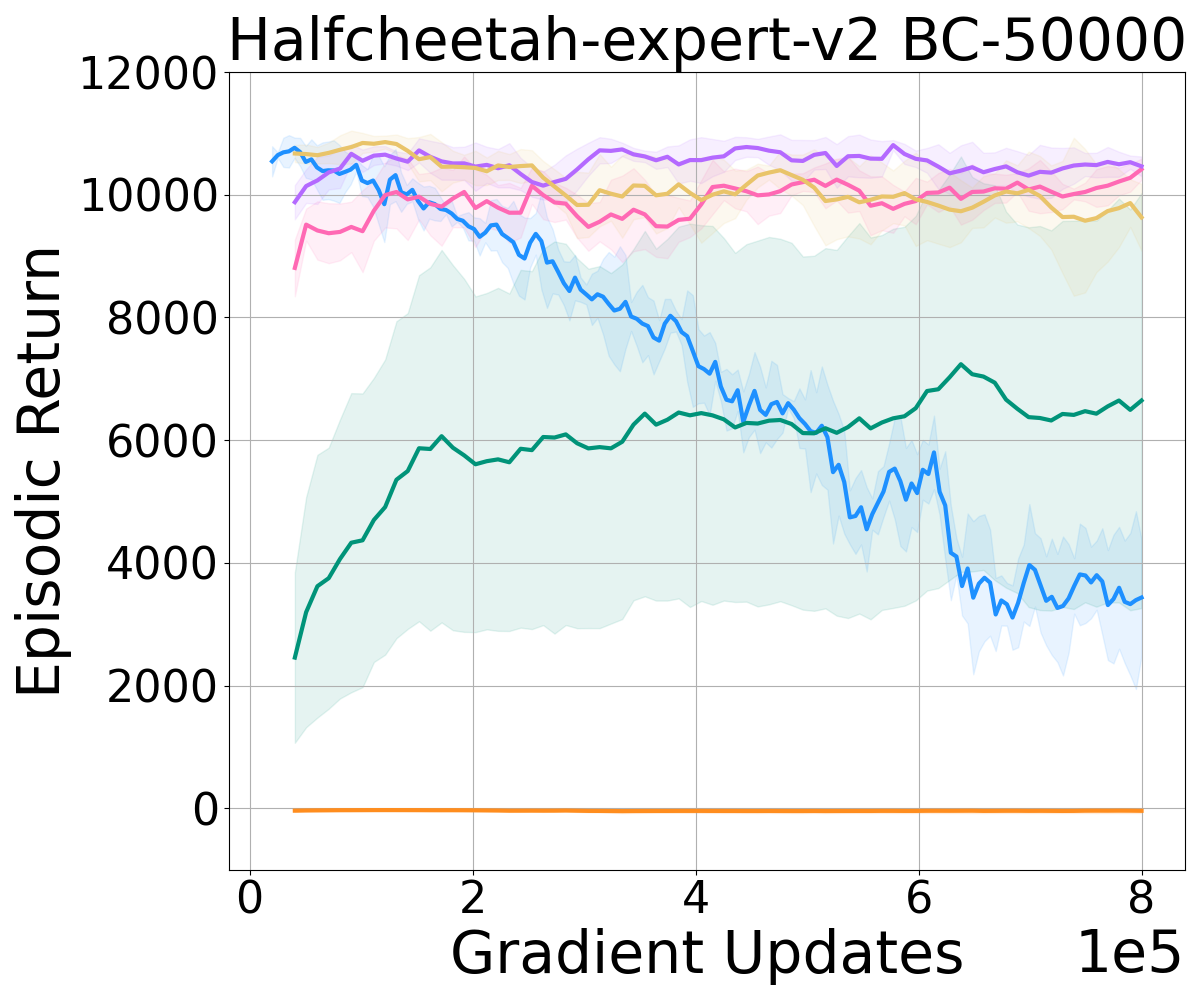}
    \caption{}
  \end{subfigure}
}
\caption{Learning curve of BC-sparse and BC-baseline for \texttt{HalfCheetah-v2} with (a) 5k and (b) 50k training samples over 800k gradient steps.}
\vspace{-10pt}
\label{fig:BC_kr_ablation}
\end{wrapfigure}

In Figure \ref{fig:BC_kr_ablation}, we conducted an ablation study where we vary the sparsity ratio and show the performance dependency on the network complexity. As we increase the training sample size from $5k$ to $50k$, we note a transition in optimal performance from a $95\%$ sparse network to a $75\%$ sparse network. This indicates the necessity of increasing network complexity with increasing training sample sizes. Consequently, for all experiments involving $5k$ and $10k$ samples, we maintain a $95\%$ sparsity, while for $50k$ and $100k$ samples, we employ a $75\%$ sparsity. Further experimental results on IQL are presented in Figure \ref{tab:vary_sparsity_datasize_performance}, which support our hypothesis on the relationship between dataset size and network sparsity in improving algorithm's performance.

%% file: files/highlight_table_new_results.tex
\definecolor{mine}{RGB}{205, 232, 248}
\newcommand{\mathcolorbox}[1]{\colorbox{mine}{\displaystyle #1}}
\newcommand{\redcolorbox}[1]{\colorbox{red!25}{\displaystyle #1}}

\begin{table*}[!]
\small
\caption{Performance improvement (without sparsity $\rightarrow$ with sparsity) of offline RL algorithms in D4RL benchmark (Expert) varying dataset size. $\pm$ captures the standard deviation across 5 seeds. \sethlcolor{mine}\hl{Performance improvements} are highlighted blue and \sethlcolor{red!25}\hl{performance losses} are highlighted red. The only performance loss is insignificant.}
\label{tab:offlineRL_baseline}
\centering
\begin{tabularx}{0.99\textwidth}{cl*3{>{\raggedright\arraybackslash}X}}
\toprule
& & HalfCheetah & Hopper & Walker2d \\
\midrule
\multirow{8}{*}{\rotatebox[origin=c]{90}{Dataset Size: 5k}} 
& \multirow{2}{*}{TD3+BC}
  & $-260.51 \pm 136.23$
  & $425.92 \pm 266.16$
  & $132.55 \pm 65.53$ \\
& 
  & $\rightarrow$ \colorbox{mine}{$494.55 \pm 796.65$}
  & $\rightarrow$ \colorbox{mine}{$1173.76 \pm 650.49$}
  & $\rightarrow$ \colorbox{mine}{$1315.61 \pm 283.20$} \\
\cmidrule{2-5}
& \multirow{2}{*}{AWAC}
  & $-158.71 \pm 89.93$
  & $410.17 \pm 116.19$
  & $117.91 \pm 39.71$ \\
&
  & $\rightarrow$ \colorbox{mine}{$3064.07 \pm 1577.04$}
  & $\rightarrow$ \colorbox{mine}{$702.99 \pm 395.07$}
  & $\rightarrow$ \colorbox{mine}{$3161.73 \pm 691.59$} \\
\cmidrule{2-5}
& \multirow{2}{*}{IQL}
  & $19.54 \pm 257.07$
  & $895.89 \pm 131.78$
  & $322.85 \pm 78.23$ \\
&
  & $\rightarrow$ \colorbox{mine}{$2454.94 \pm 1002.59$}
  & $\rightarrow$ \colorbox{mine}{$2075.30 \pm 749.97$}
  & $\rightarrow$ \colorbox{mine}{$3988.02 \pm 814.09$} \\
\midrule\midrule

\multirow{10}{*}{\rotatebox[origin=c]{90}{Dataset Size: 10k}}  
& \multirow{2}{*}{TD3+BC}
  & $199.03 \pm 168.51$
  & $1127.84 \pm 477.30$
  & $192.66 \pm 48.67$ \\
&
  & $\rightarrow$ \colorbox{mine}{$1038.07 \pm 1262.93$}
  & $\rightarrow$ \colorbox{mine}{$1860.95 \pm 800.67$}
  & $\rightarrow$ \colorbox{mine}{$3121.36 \pm 285.98$} \\
\cmidrule{2-5}
& \multirow{2}{*}{AWAC}
  & $-484.61 \pm 56.03$
  & $203.83 \pm 39.02$
  & $44.12 \pm 32.77$ \\
&
  & $\rightarrow$ \colorbox{mine}{$8735.16 \pm 627.81$}
  & $\rightarrow$ \colorbox{mine}{$2142.28 \pm 855.29$}
  & $\rightarrow$ \colorbox{mine}{$4596.78 \pm 262.41$} \\
\cmidrule{2-5}
& \multirow{2}{*}{IQL}
  & $218.04 \pm 67.31$
  & $907.67 \pm 252.13$
  & $224.13 \pm 62.11$ \\
&
  & $\rightarrow$ \colorbox{mine}{$8521.96 \pm 865.68$}
  & $\rightarrow$ \colorbox{mine}{$2120.03 \pm 404.96$}
  & $\rightarrow$ \colorbox{mine}{$4927.65 \pm 47.62$} \\
\midrule\midrule


\bottomrule
\end{tabularx}
\end{table*}

%% file: files/07_conclusion.tex
\section{Discussion}


We explore the use of small datasets within the realm of offline RL for continuous control tasks. While many standard offline RL benchmarks involve datasets containing well over a million data points, it is important to recognize that numerous applications operate with much smaller datasets. Our primary focus is on reducing the overfitting of offline RL algorithms when dealing with smaller training datasets. 

To tackle this challenge, we introduce a regularization method centred on the concept of sparsity. Our experiments demonstrate that incorporating sparsity in neural networks can significantly enhance performance efficiency across a variety of algorithms, datasets, and tasks. Additionally, we perform an examination of various regularization methods. Our results highlight that sparsity, through targeted regularization of the parameter space, proves to be a more effective approach for achieving successful learning with limited datasets.



%% file: files/08_appendix.tex
\newpage
\section*{Appendix}

\paragraph{Why not use a smaller neural network?}
The impact of training a \emph{smaller network} and training \emph{a subnetwork of a larger network} is different. A smaller network inherently has fewer degrees of freedom because it has fewer neurons and parameters, which reduces its ability to model complex tasks. On the other hand, when training a subnetwork within a larger network, you select a specific subspace of parameters better suited to the task at hand, while still leveraging the full architecture and its larger representational capacity. This approach targets a more relevant portion of the parameter space, allowing the model to focus on the most important connections and regularize the training process. While the number of active parameters is smaller, the remaining connections in the full network still enable the model to learn complex tasks.
We present empirical results in Table \ref{tab:iql_performance_smaller_net} to support our hypothesis. While we observe an improvement in performance over the IQL baseline (hidden-dimension=256), reducing the model size (e.g., using hidden dimensions of 64 and 128 for both actor and critic networks) proves unreliable and does not deliver consistent performance compared to IQL-Sparse-Reg.

\begin{table}[h!]
\small
\centering
\begin{tabular}{|c|c|c|c|c|}
\hline
\textbf{Environment}    & \textbf{IQL-baseline}      & \textbf{IQL HD=64}       & \textbf{IQL HD=128}      & \textbf{IQL-Sparse-Reg}         \\ \hline
HalfCheetah-Expert     & $19.54 \pm 257.07$         & $181.22 \pm 593.25$         & $-403.65 \pm 66.39$         & \textbf{2454.94 ± 1002.59}      \\ \hline
Hopper-Expert          & $895.89 \pm 131.78$        & $740.92 \pm 285.88$         & $383.31 \pm 74.44$          & \textbf{2075.3 ± 749.97}         \\ \hline
\end{tabular}
\caption{Performance comparison of Sparse-Reg with varying dense network size.}
\label{tab:iql_performance_smaller_net}
\end{table}

\paragraph{Performance at Various Sparsity Levels for Different Dataset Sizes:}
We present an empirical study in Table \ref{tab:vary_sparsity_datasize_performance} that shows how performance varies as we adjust sparsity and dataset size. The experiment is conducted using IQL on the \texttt{HalfCheetah}-Expert task. This supports our hypothesis and aligns with previously observed results in Figure \ref{tab:iql_performance_smaller_net}, where we find that as the number of training samples increases, more trainable parameters are needed. For dataset sizes of 50k and 100k, the best performance is achieved with 75\% sparsity.
\begin{table}[h!]
\small
\centering
\begin{tabularx}{0.99\textwidth}{l*5{>{\raggedright\arraybackslash}X}}
\hline
\textbf{Dataset Size} & \textbf{Sparsity 99\%} & \textbf{Sparsity 95\%} & \textbf{Sparsity 75\%} & \textbf{Sparsity 50\%} & \textbf{Sparsity 20\%} \\ \hline
\multirow{2}{*}{5K}  & $108.58$      & \textbf{2454}     & $-322.96$      & $-17.16$        & $-182.57$      \\ 
                     & $\pm 197.91$   & \textbf{± 1002.59} & $\pm 82.27$    & $\pm 194.26$    & $\pm 67.33$    \\ \hline
\multirow{2}{*}{10K} & $289.55$      & \textbf{8521.96}  & $92.35$        & $-280.0$        & $-280.46$      \\
                     & $\pm 226.44$   & \textbf{± 865.68} & $\pm 99.9$     & $\pm 99.16$     & $\pm 148.1$    \\ \hline
\multirow{2}{*}{50K} & $52.01$       & $9939.25$         & \textbf{10644.76} & $8538.45$       & $6685.56$      \\
                     & $\pm 443.25$   & $\pm 162.37$      & \textbf{± 123.53} & $\pm 1444.19$   & $\pm 673.57$   \\ \hline
\multirow{2}{*}{100K} & $544.22$      & $10574.44$        & \textbf{11116.75} & $10987.68$      & $10728.38$     \\
                     & $\pm 994.11$   & $\pm 483.28$      & \textbf{± 65.37}     & $\pm 118.84$    & $\pm 196.58$   \\ \hline
\end{tabularx}
\caption{Performance of IQL-Sparse-Reg at various sparsity levels for different data sizes for \texttt{HalfCheetah}-Expert task.}
\label{tab:vary_sparsity_datasize_performance}
\end{table}

\paragraph{Performance for Medium dataset:} We present detailed benchmark on \emph{Medium} dataset here in Table \ref{tab:offlineRL_baseline_medium}.

\begin{figure}
\centering
\scalebox{0.7}{
   \includegraphics[width=0.33\linewidth]{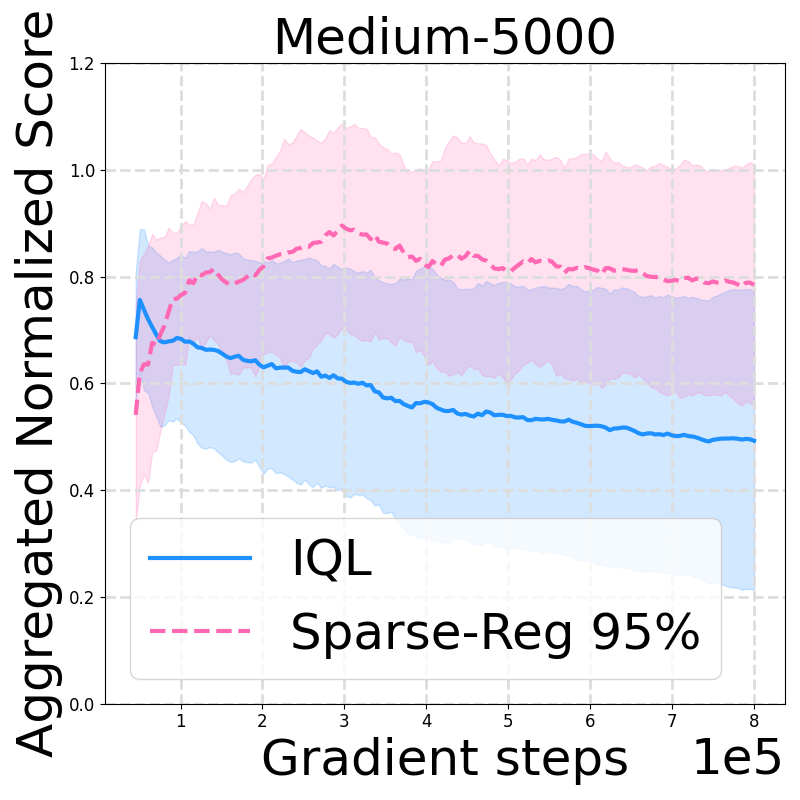}
  \includegraphics[width=0.33\linewidth]{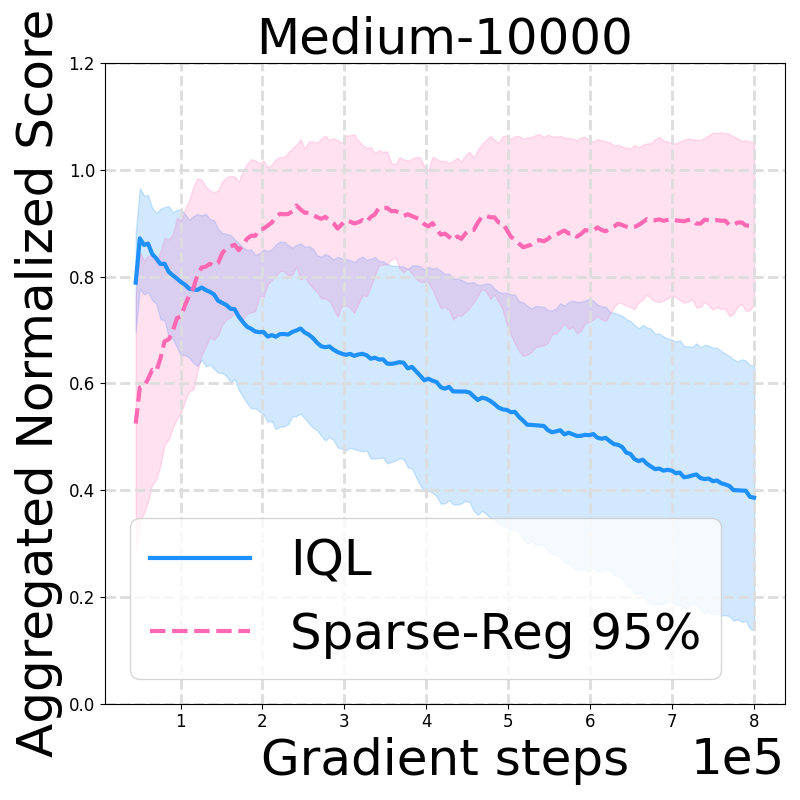}
\includegraphics[width=0.33\linewidth]{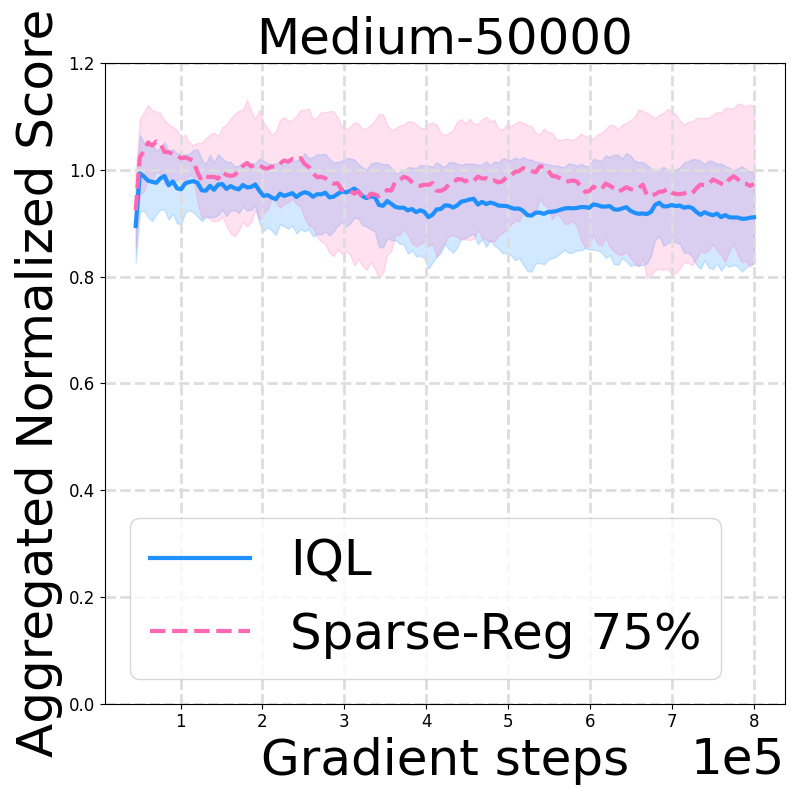}
\includegraphics[width=0.33\linewidth]{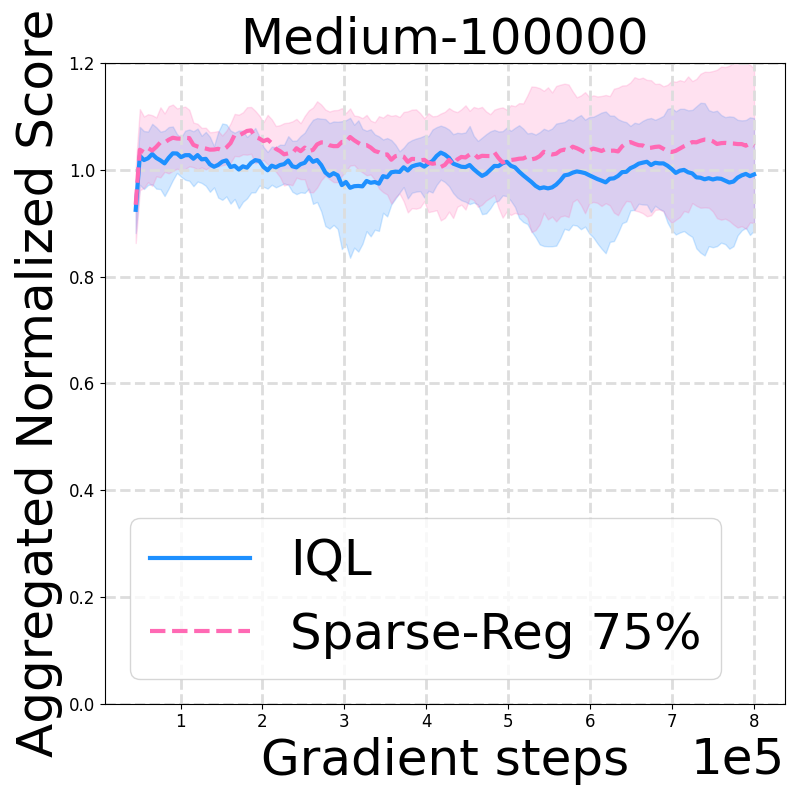}
 } 
  \vspace{-6pt}
  \caption{Aggregated normalized performance of IQL and IQL-Sparse-Reg across \texttt{HalfCheetah}, \texttt{Hopper}, and \texttt{Walker2d} environments with varying training sample sizes: 5k, 10k, 50k, and 100k on D4RL-\emph{Medium} dataset.}
\label{fig:agg_IQL_medium_mujoco}
\end{figure}

\input{files/highlight_medium}


\paragraph{Aggregated Normalized Performance:}
We present additional results on the learning curve of aggregated normalized performance for the D4RL \cite{D4RL} \emph{Medium} dataset in Figure \ref{fig:agg_IQL_medium_mujoco}. These results further demonstrate that applying sparse regularization enhances the sampling complexity of offline RL algorithms.

\paragraph{Performance for Mixed Dataset:} Offline datasets can originate from a mixture of different distributions, which can introduce variability in the data. To account for this in Table \ref{tab:iql_mixed_dataset}, we present the performance of IQL on two environments, \texttt{HalfCheetah} and \texttt{Hopper}, using datasets of 5k and 10k samples on mixed datasets: Medium-Replay and Expert-Replay.

\begin{table}[h!]
\centering
\begin{tabular}{|c|c|c|c|c|}
\hline
\textbf{Environment} & \textbf{Dataset size} & \textbf{Mixed-data}   & \textbf{IQL-baseline}      & \textbf{IQL-Sparse-Reg}         \\ \hline
\multirow{4}{*}{HalfCheetah}  & 5k  & Medium-Replay         & 85.97 ± 7.68               & \textbf{147.97 ± 8.55}          \\ \cline{2-5}
                              &      & Expert-Replay        & 28.2 ± 5.14                & \textbf{82.91 ± 28.97}          \\ \cline{2-5}
                              & 10k & Medium-Replay         & 68.47 ± 11.97              & \textbf{122.73 ± 8.25}          \\ \cline{2-5}
                              &      & Expert-Replay        & 21.3 ± 2.38                & \textbf{92.49 ± 12.19}          \\ \hline
\multirow{4}{*}{Hopper}       & 5k  & Medium-Replay         & 85.97 ± 7.68               & \textbf{147.97 ± 8.55}          \\ \cline{2-5}
                              &      & Expert-Replay        & 28.2 ± 5.14                & \textbf{82.91 ± 28.97}          \\ \cline{2-5}
                              & 10k & Medium-Replay         & 68.47 ± 11.97              & \textbf{122.73 ± 8.25}          \\ \cline{2-5}
                              &      & Expert-Replay        & 21.3 ± 2.38                & \textbf{92.49 ± 12.19}          \\ \hline
\end{tabular}
\caption{Performance comparison of IQL-baseline and IQL-Sparse-Reg on Mixed dataset.}
\label{tab:iql_mixed_dataset}
\end{table}

\begin{figure*}[htb!]
\small
\centering

\begin{subfigure}{\textwidth}
\centering
\includegraphics[width=0.22\linewidth]{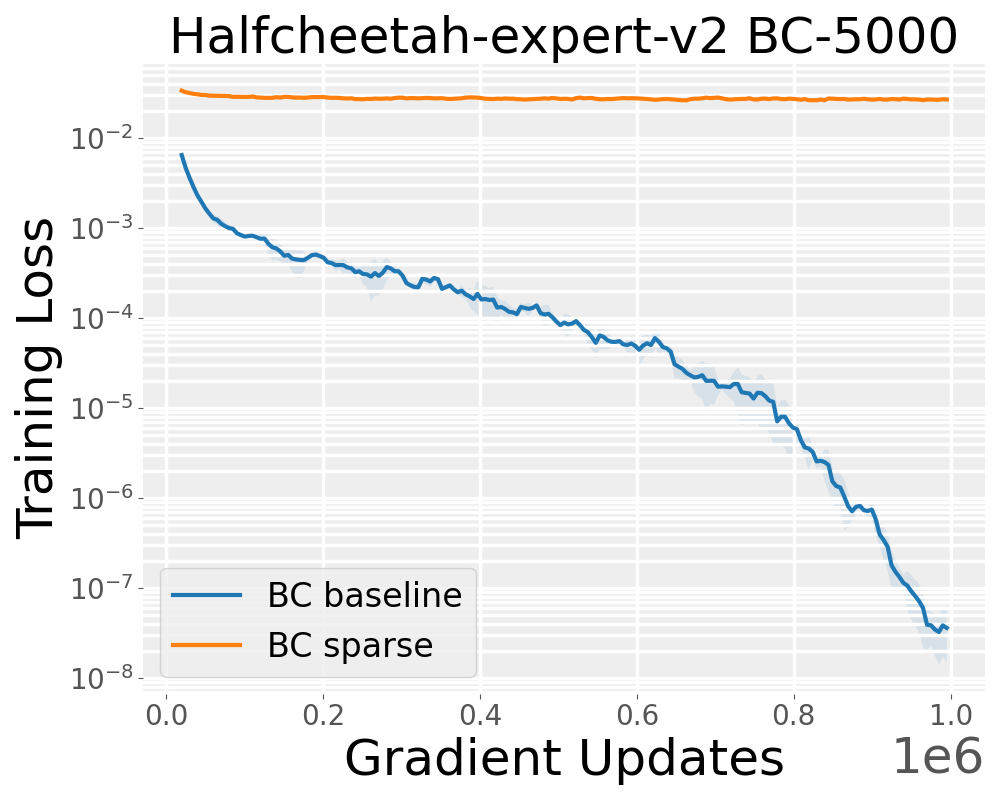}
\includegraphics[width=0.22\linewidth]{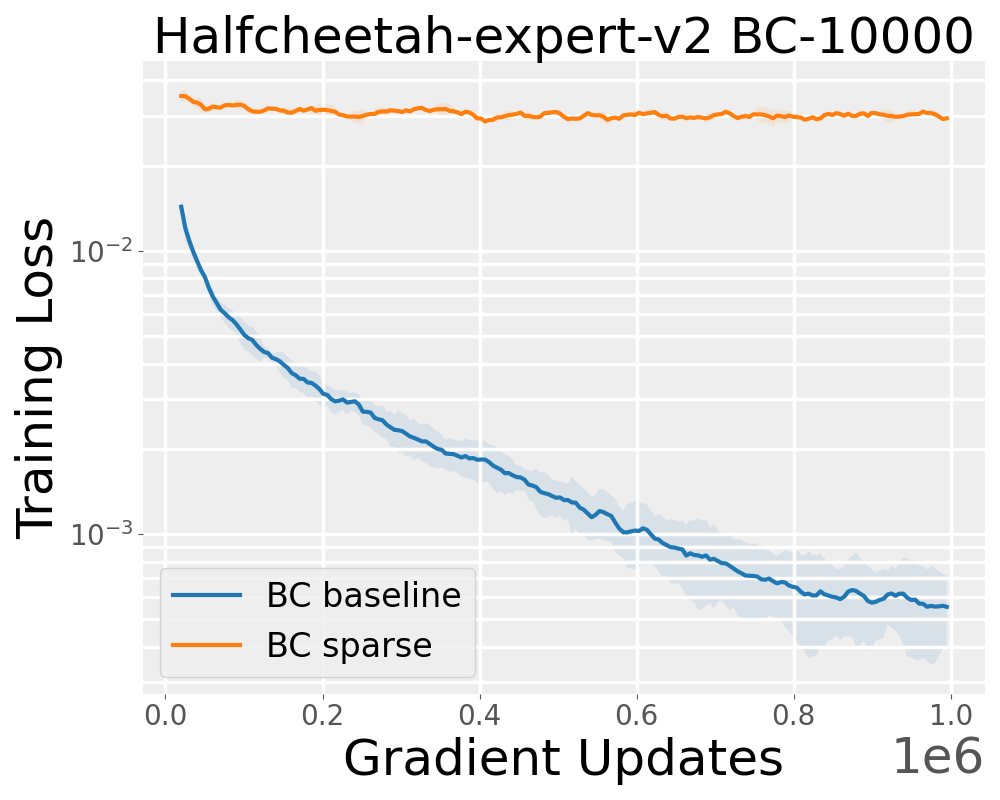}
\includegraphics[width=0.22\linewidth]{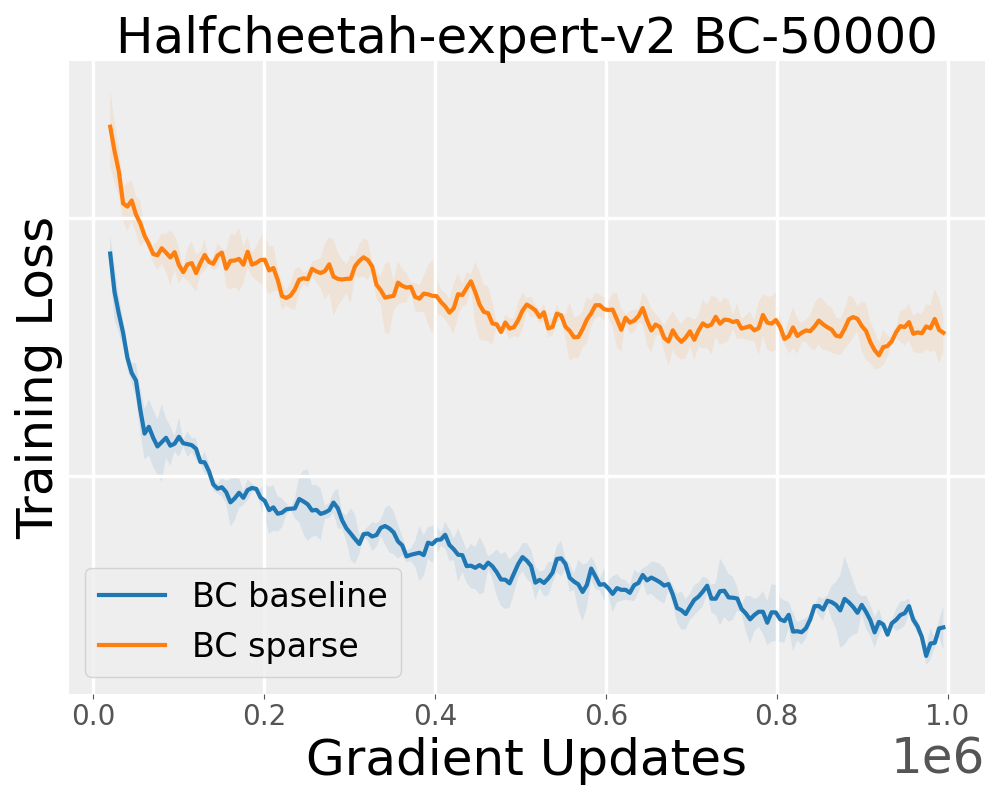}
\includegraphics[width=0.22\linewidth]{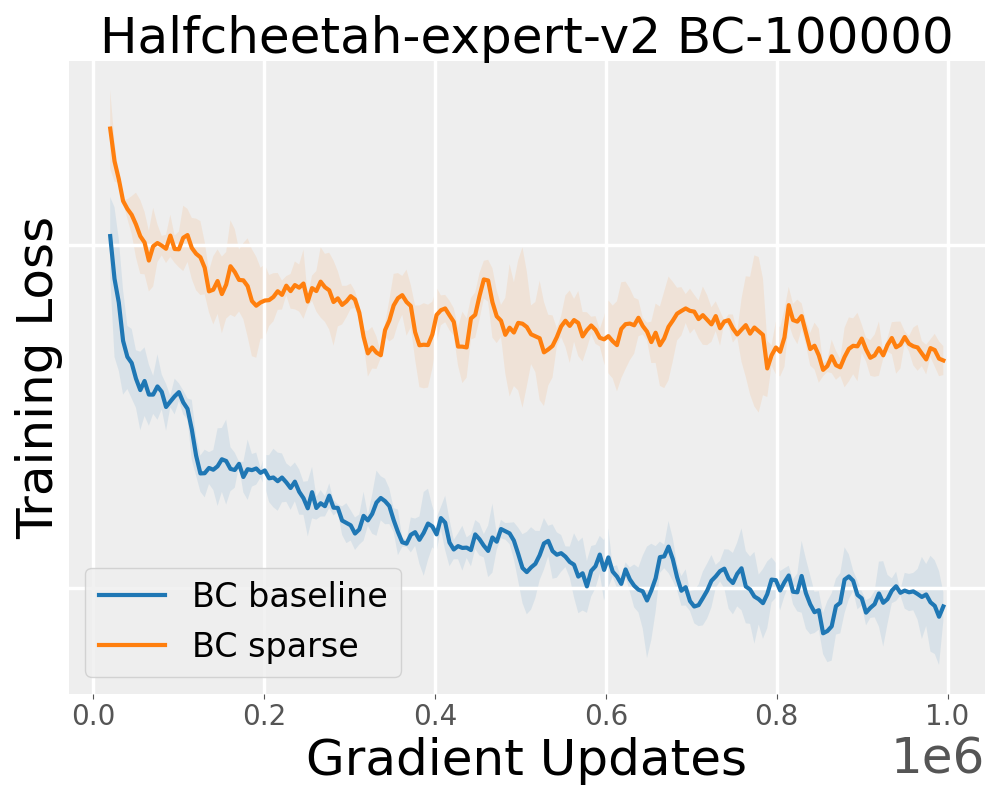}
\caption{\texttt{HalfCheetah-v2} trained with Expert dataset}
\end{subfigure}

\vspace{0.5em}

\begin{subfigure}{\textwidth}
\centering
\includegraphics[width=0.22\linewidth]{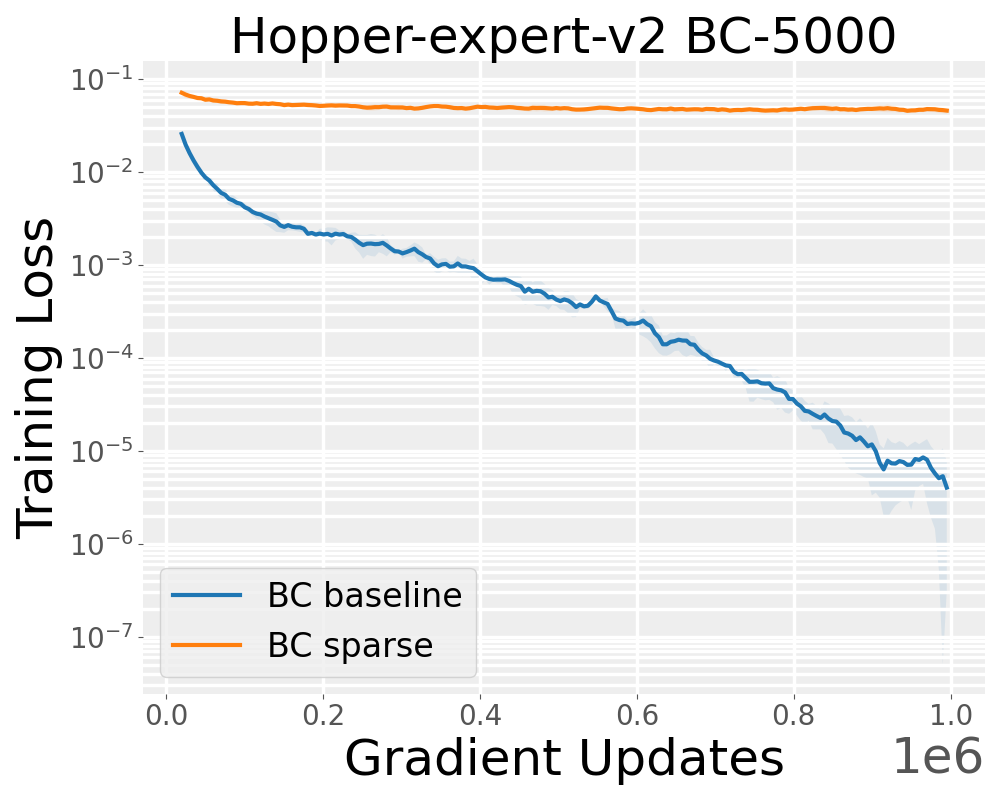}
\includegraphics[width=0.22\linewidth]{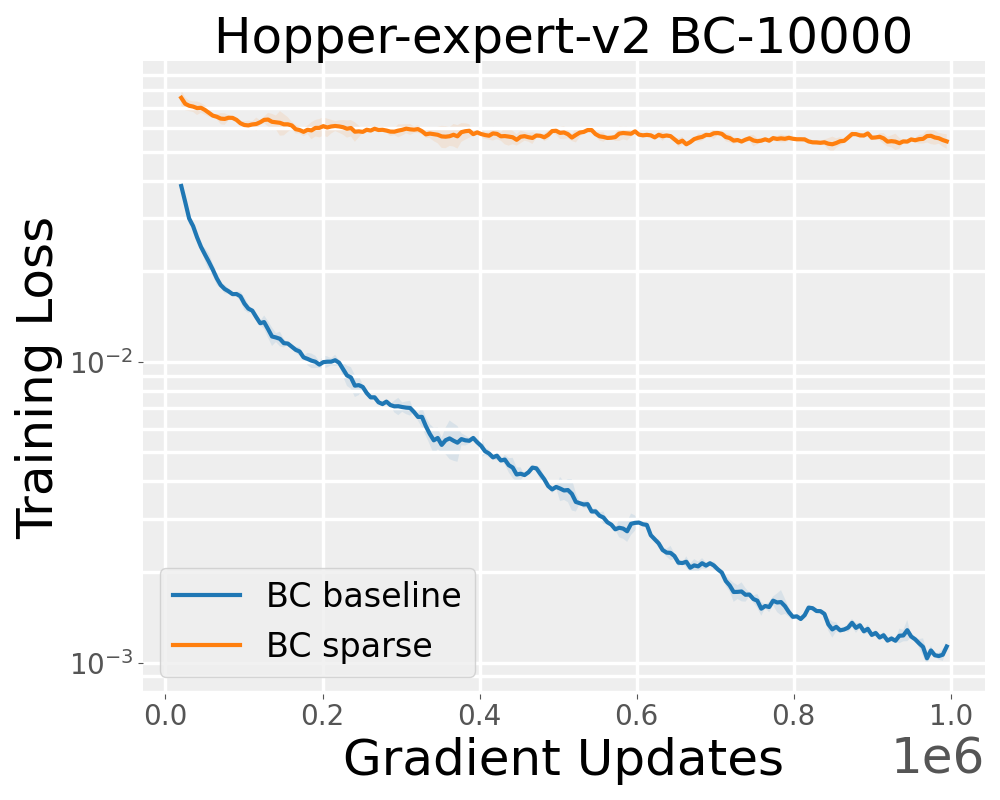}
\includegraphics[width=0.22\linewidth]{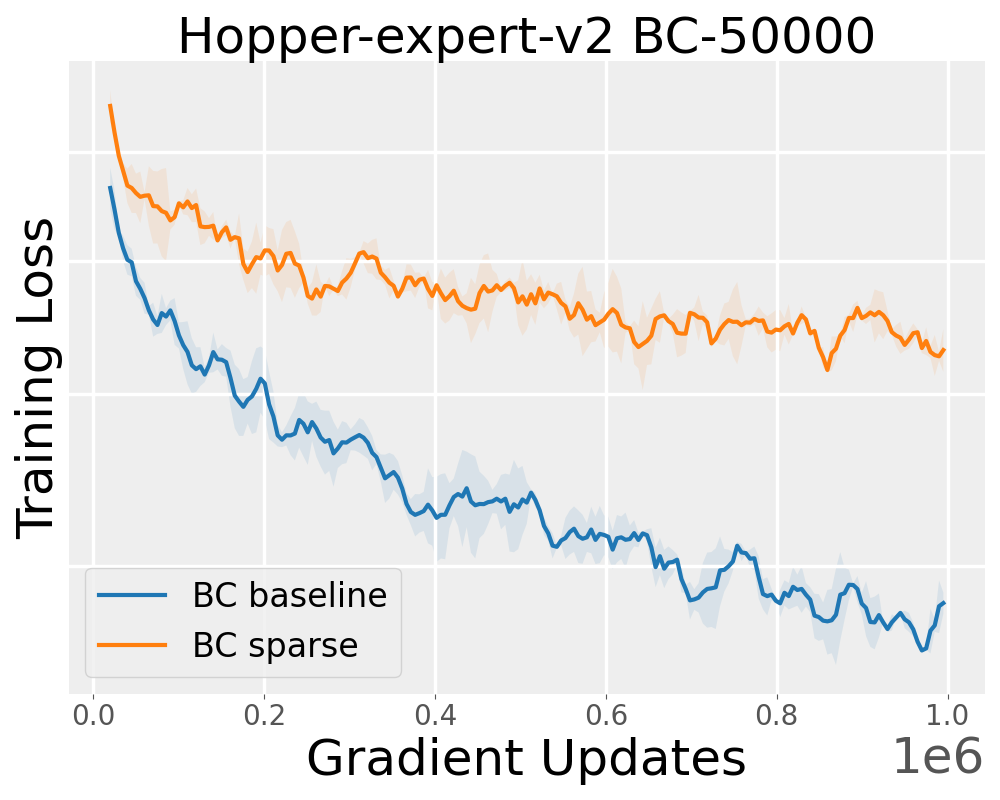}
\includegraphics[width=0.22\linewidth]{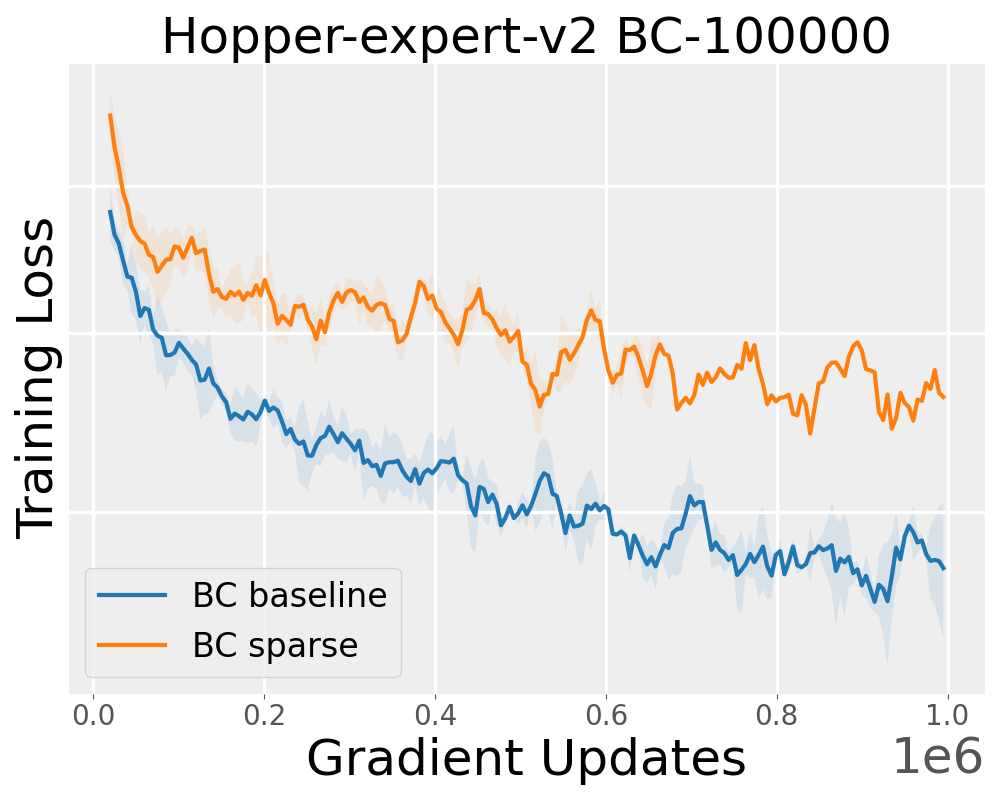}
\caption{\texttt{Hopper-v2} trained with Expert dataset}
\end{subfigure}

\vspace{0.5em}

\begin{subfigure}{\textwidth}
\centering
\includegraphics[width=0.22\linewidth]{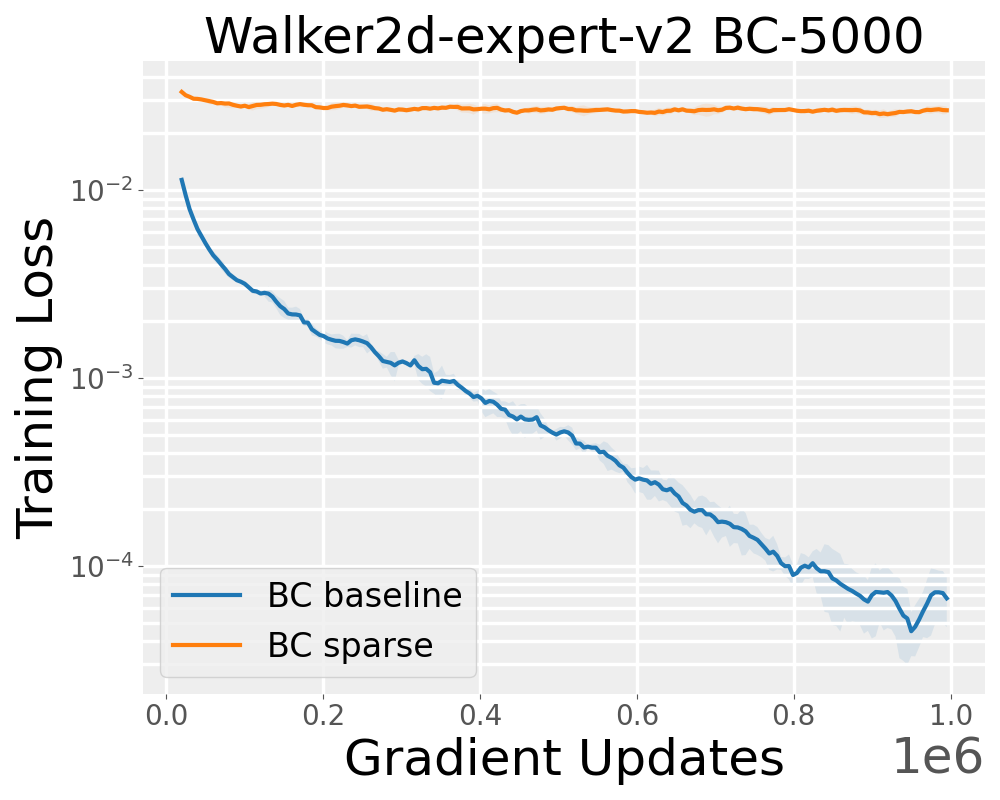}
\includegraphics[width=0.22\linewidth]{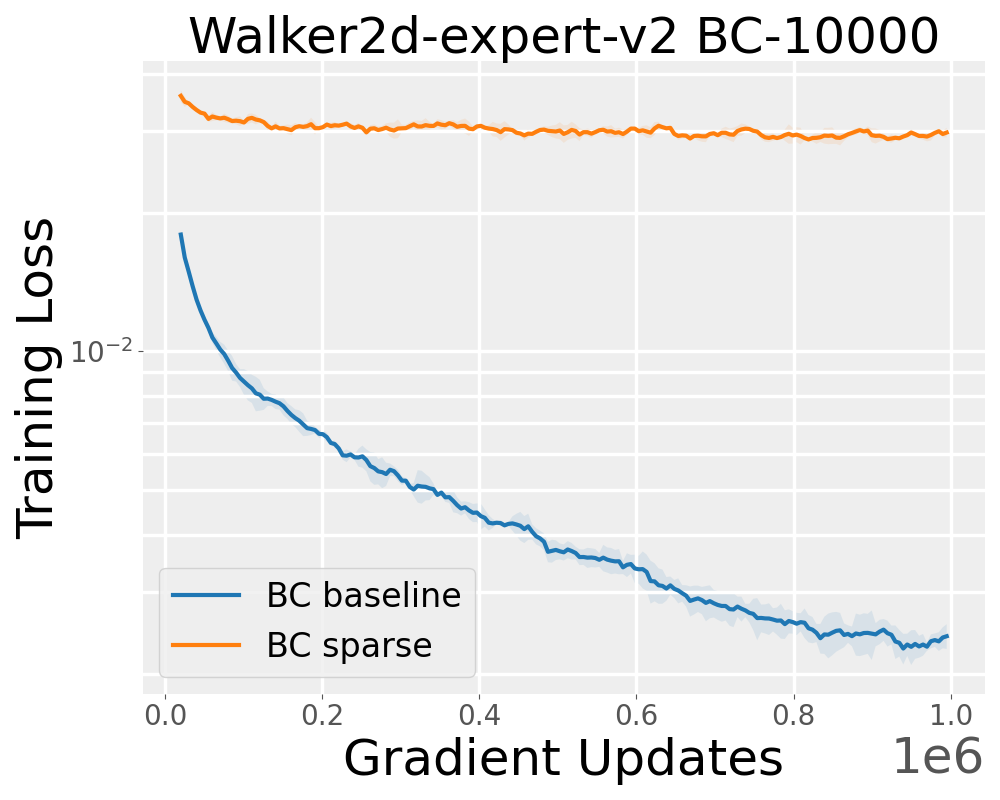}
\includegraphics[width=0.22\linewidth]{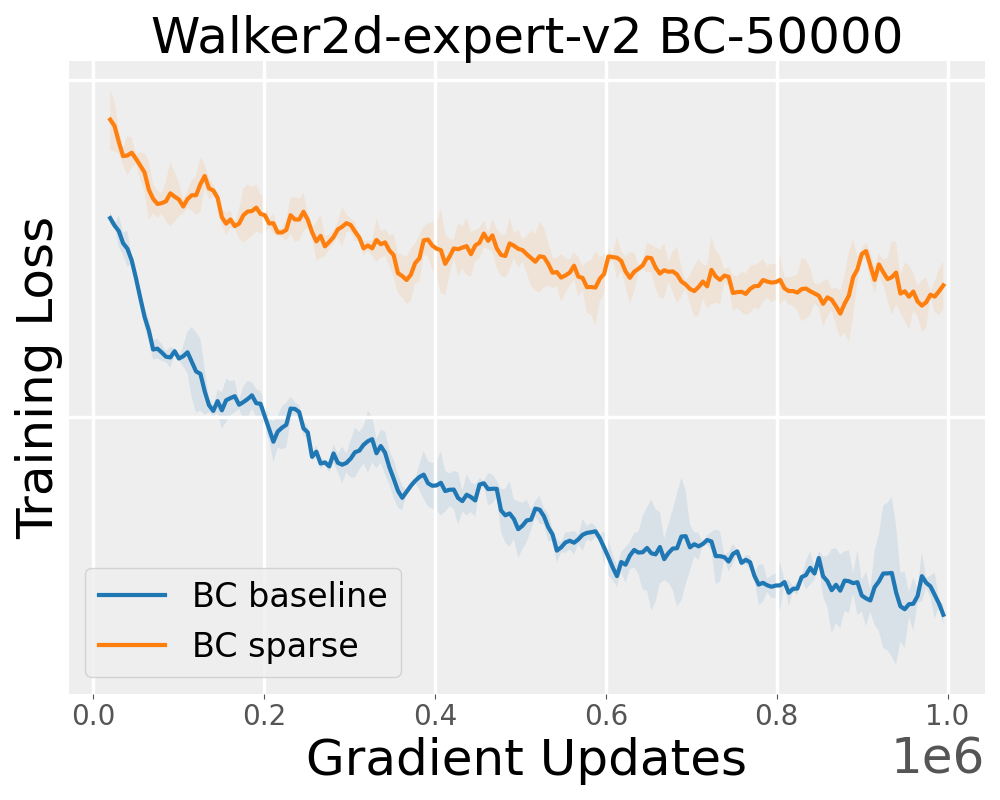}
\includegraphics[width=0.22\linewidth]{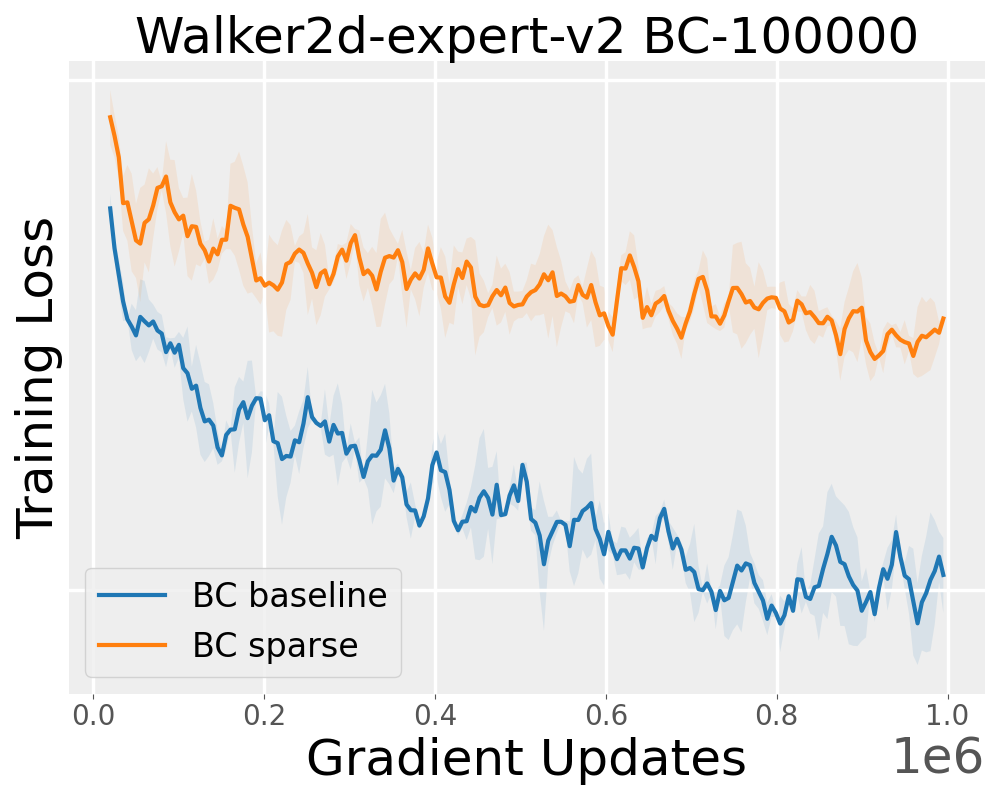}
\caption{\texttt{Walker2d-v2} trained with Expert dataset}
\end{subfigure}

\vspace{-6pt}
\caption{Compare the MSE loss of behaviour-action prediction on the \textbf{training dataset} with varying training sample size between BC-baseline (\textcolor{cyan}{blue}) and sparse regularized BC (\textcolor{orange}{orange}) over 1 million training steps.}
\label{fig:training_bc_loss}
\end{figure*}

\begin{figure*}[htb!]
\centering

\begin{subfigure}{\textwidth}
\centering
\includegraphics[width=0.18\linewidth]{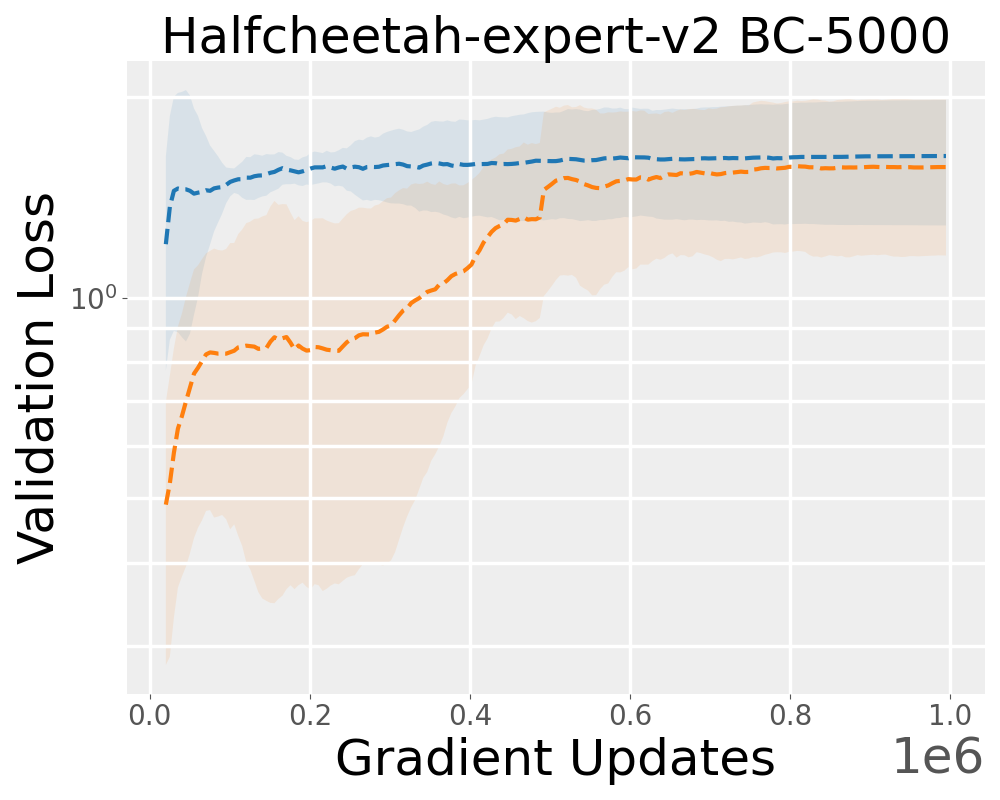}
\includegraphics[width=0.18\linewidth]{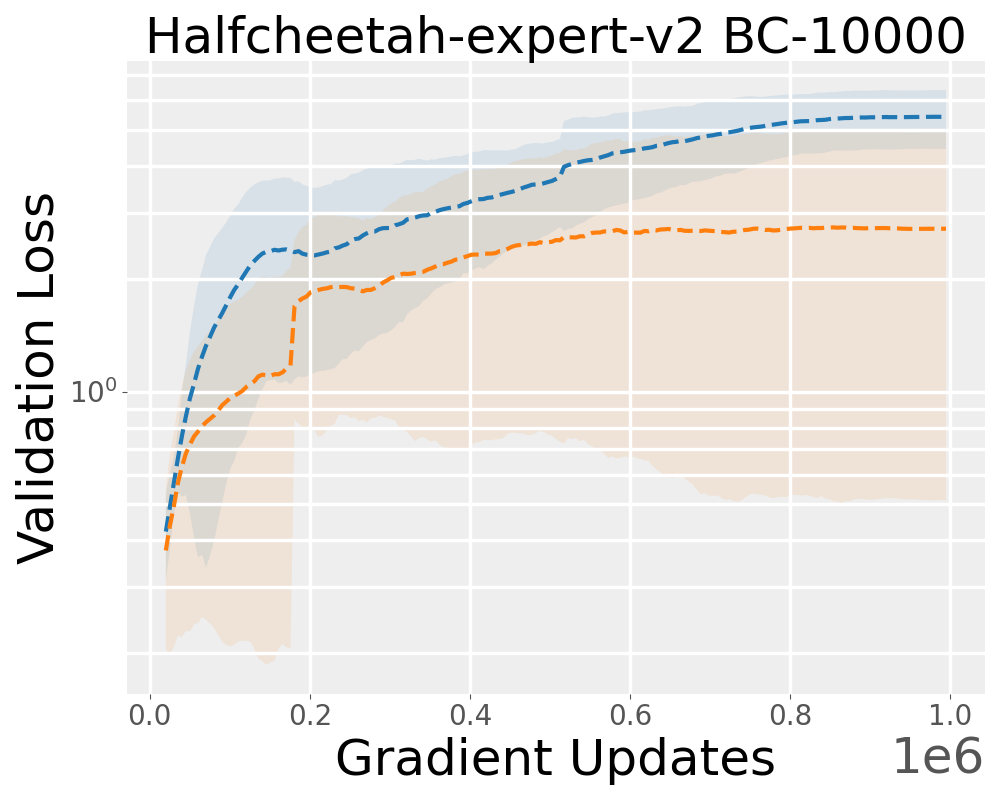}
\includegraphics[width=0.18\linewidth]{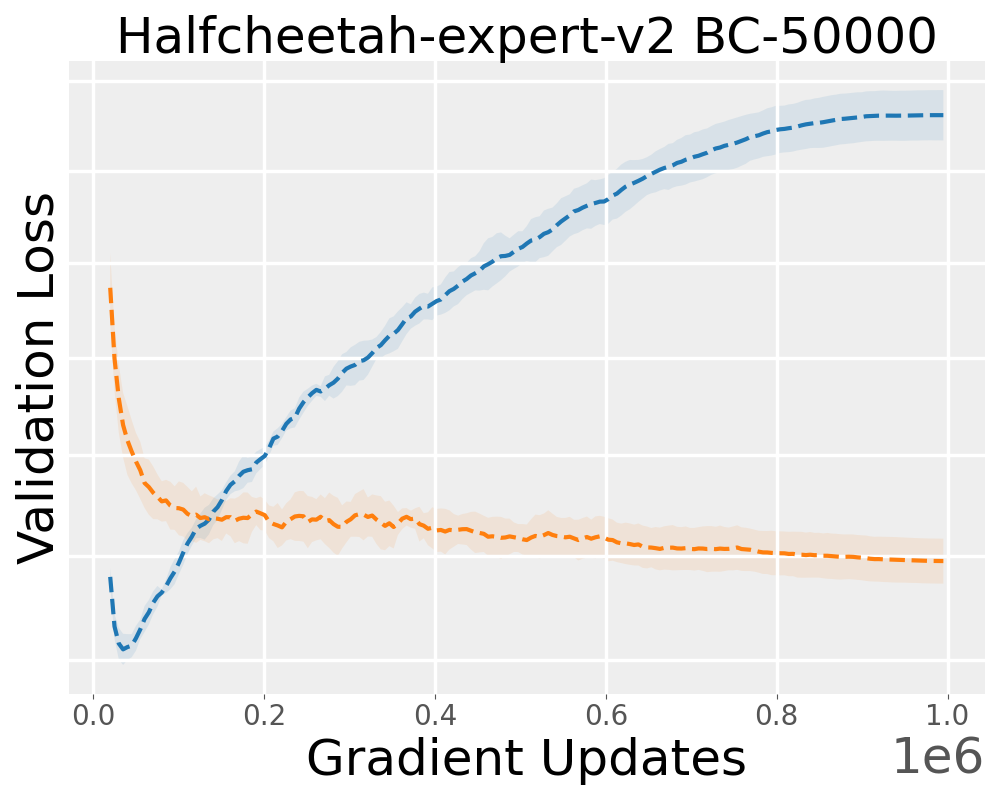}
\includegraphics[width=0.18\linewidth]{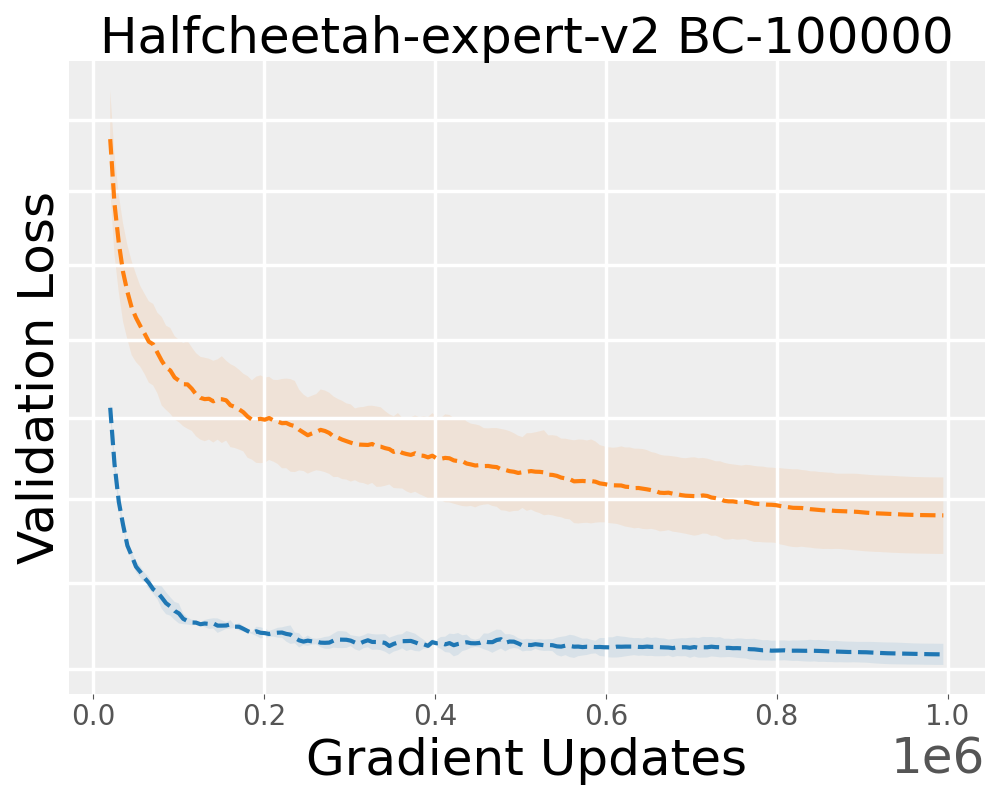}
\caption{\texttt{HalfCheetah-v2} trained with Expert dataset}
\end{subfigure}

\vspace{0.5em}

\begin{subfigure}{\textwidth}
\centering
\includegraphics[width=0.18\linewidth]{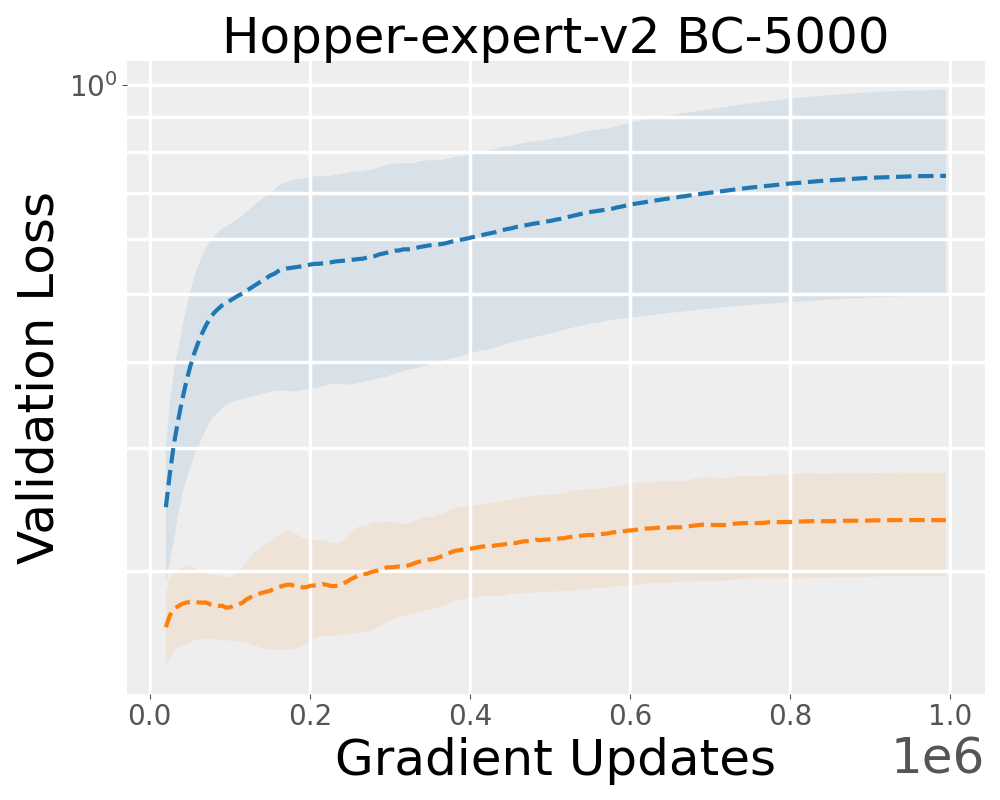}
\includegraphics[width=0.18\linewidth]{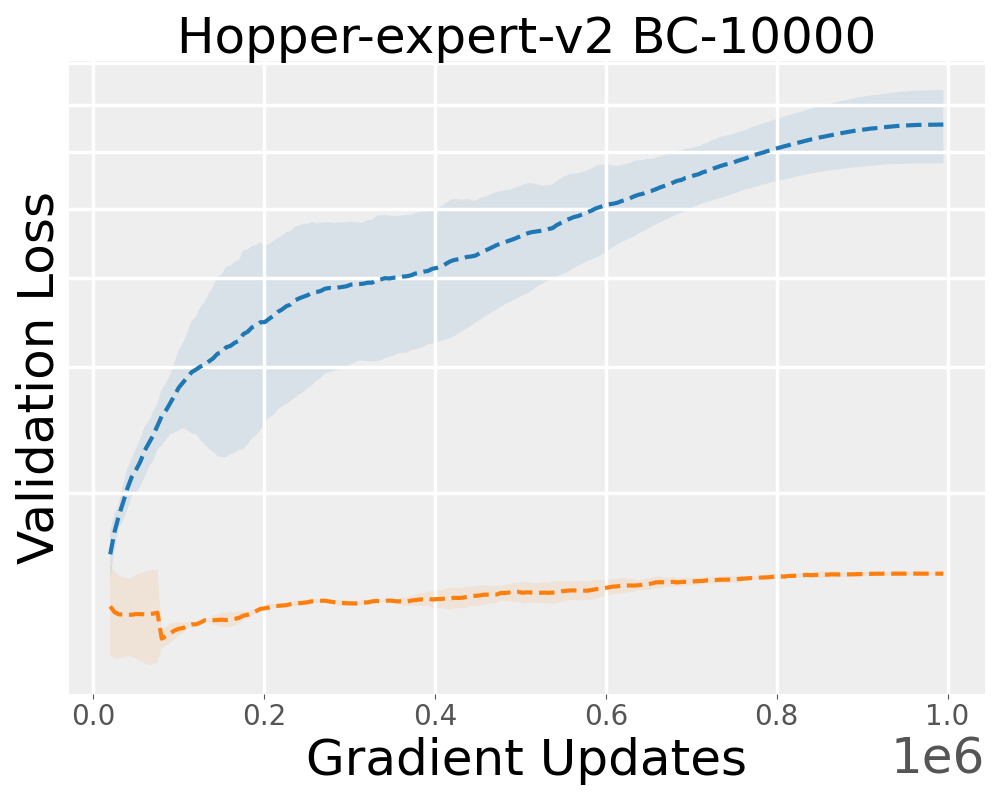}
\includegraphics[width=0.18\linewidth]{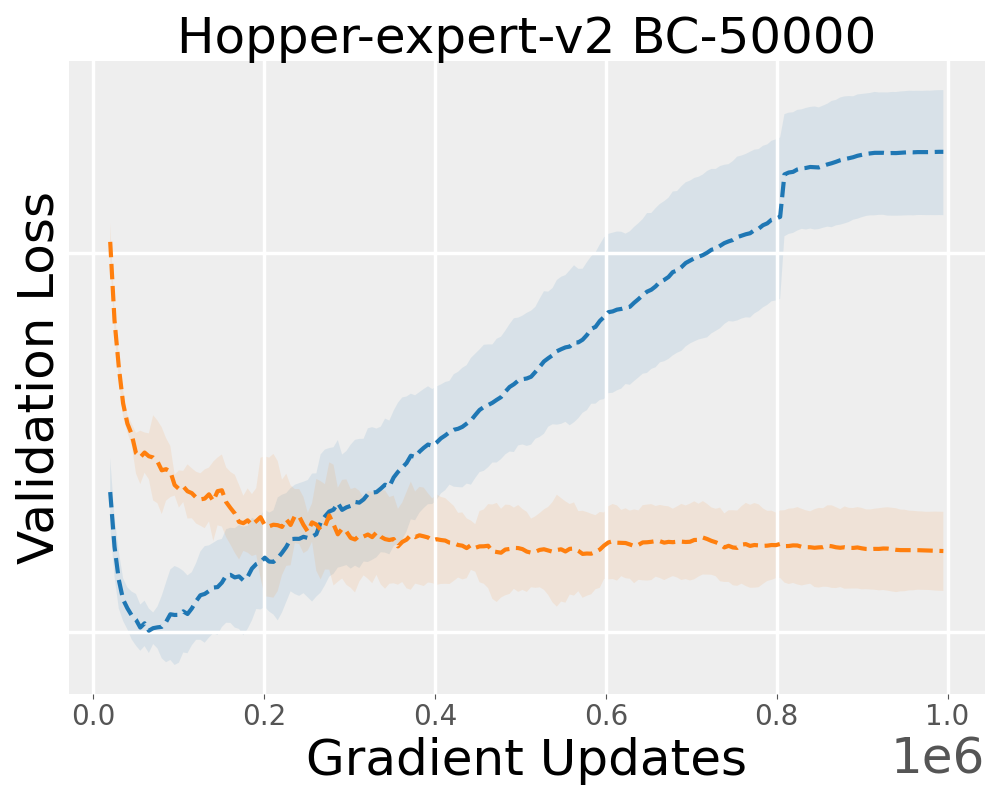}
\includegraphics[width=0.18\linewidth]{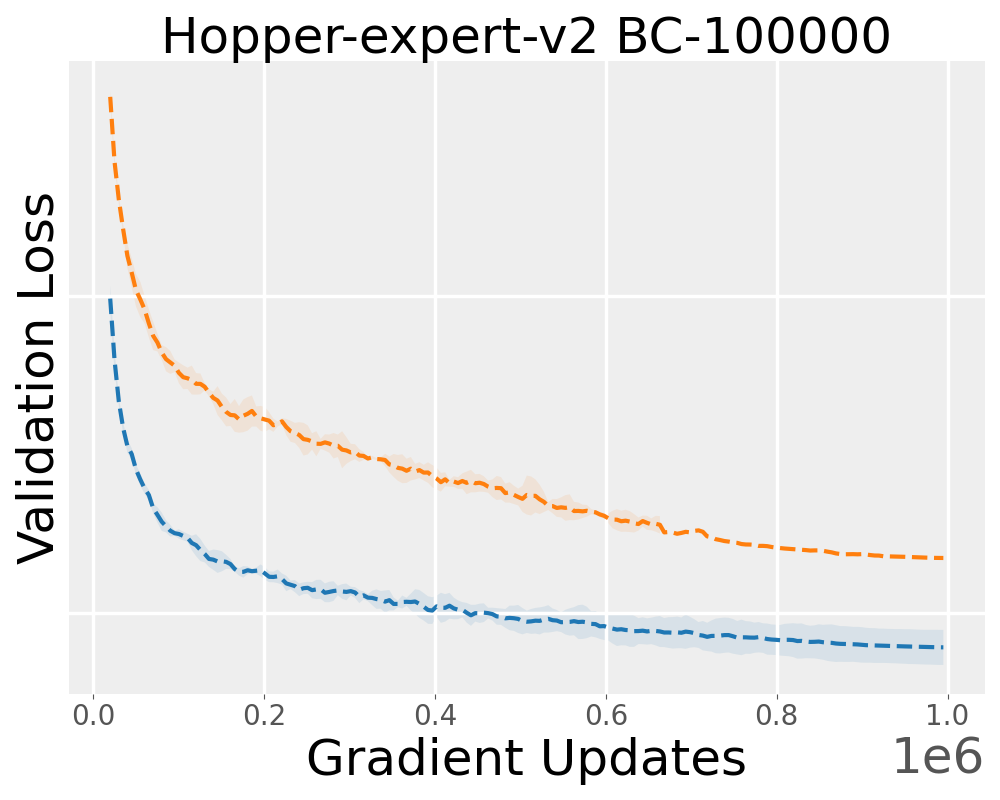}
\caption{\texttt{Hopper-v2} trained with Expert dataset}
\end{subfigure}

\vspace{0.5em}

\begin{subfigure}{\textwidth}
\centering
\includegraphics[width=0.18\linewidth]{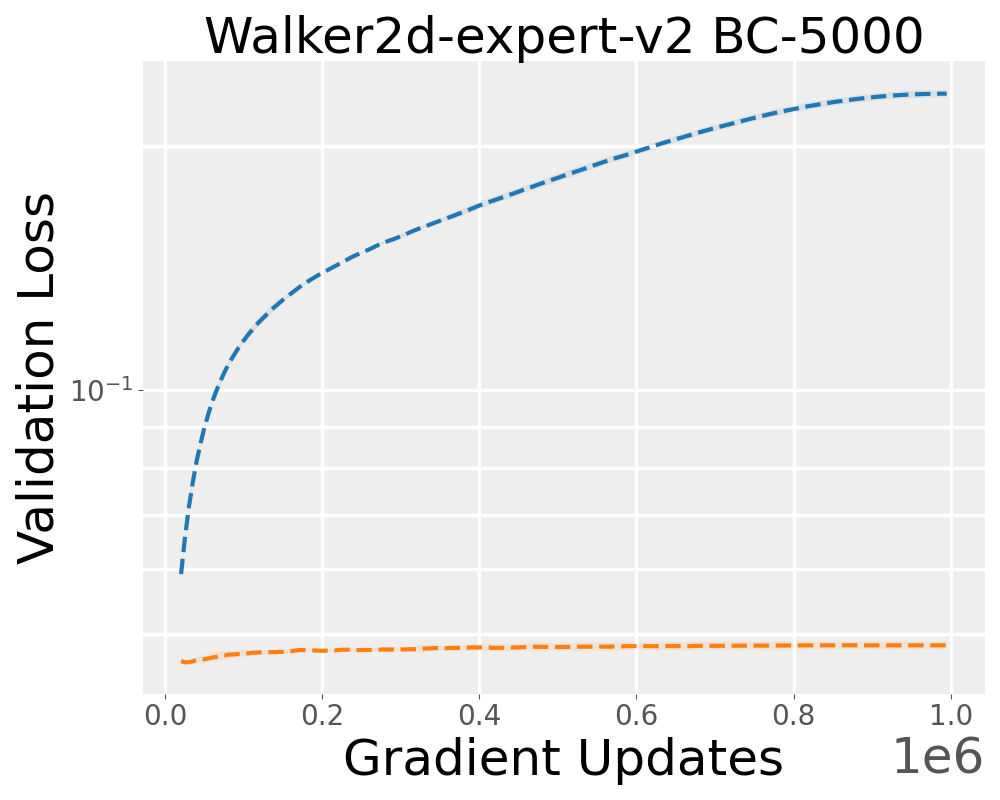}
\includegraphics[width=0.18\linewidth]{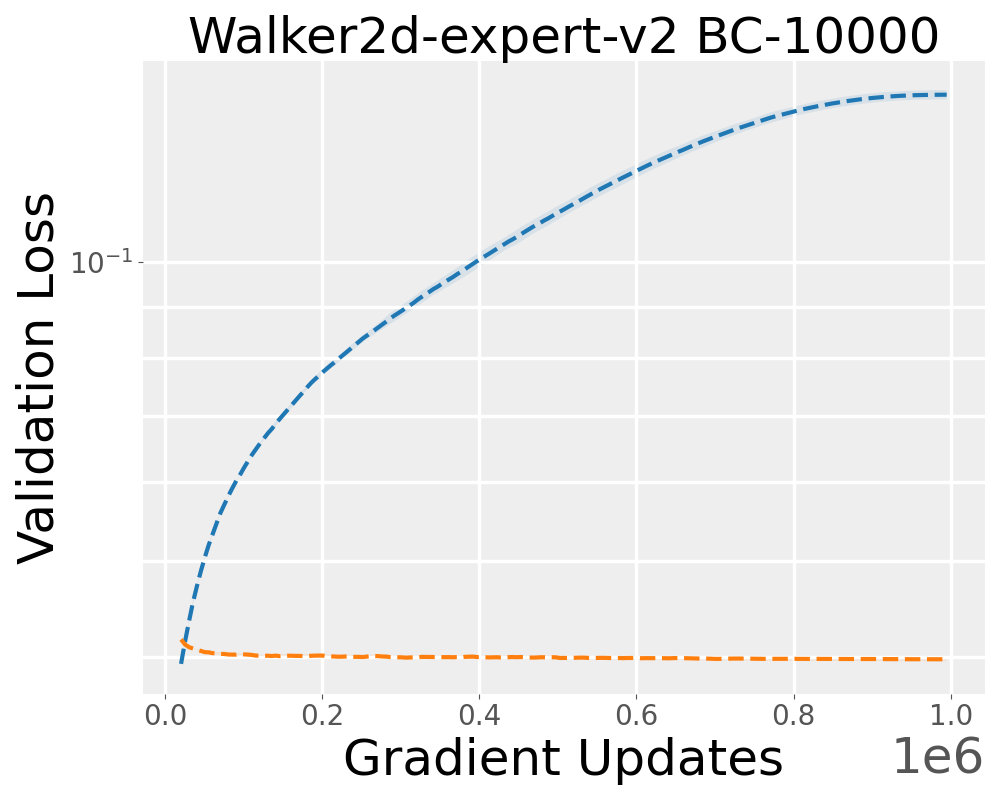}
\includegraphics[width=0.18\linewidth]{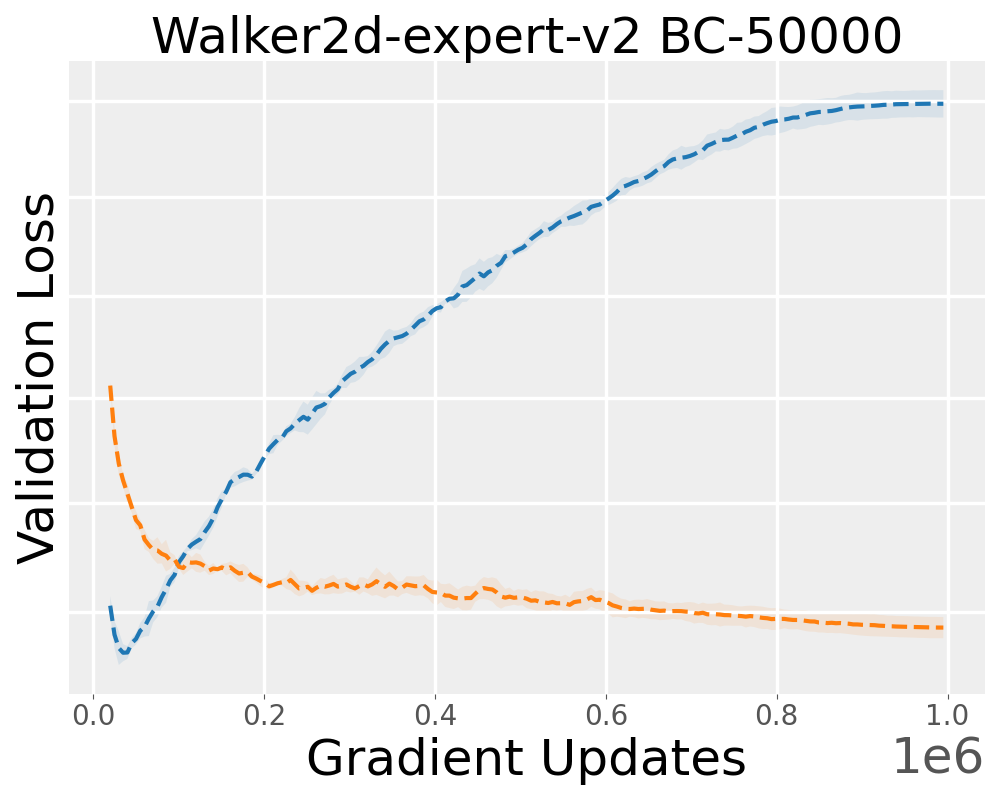}
\includegraphics[width=0.18\linewidth]{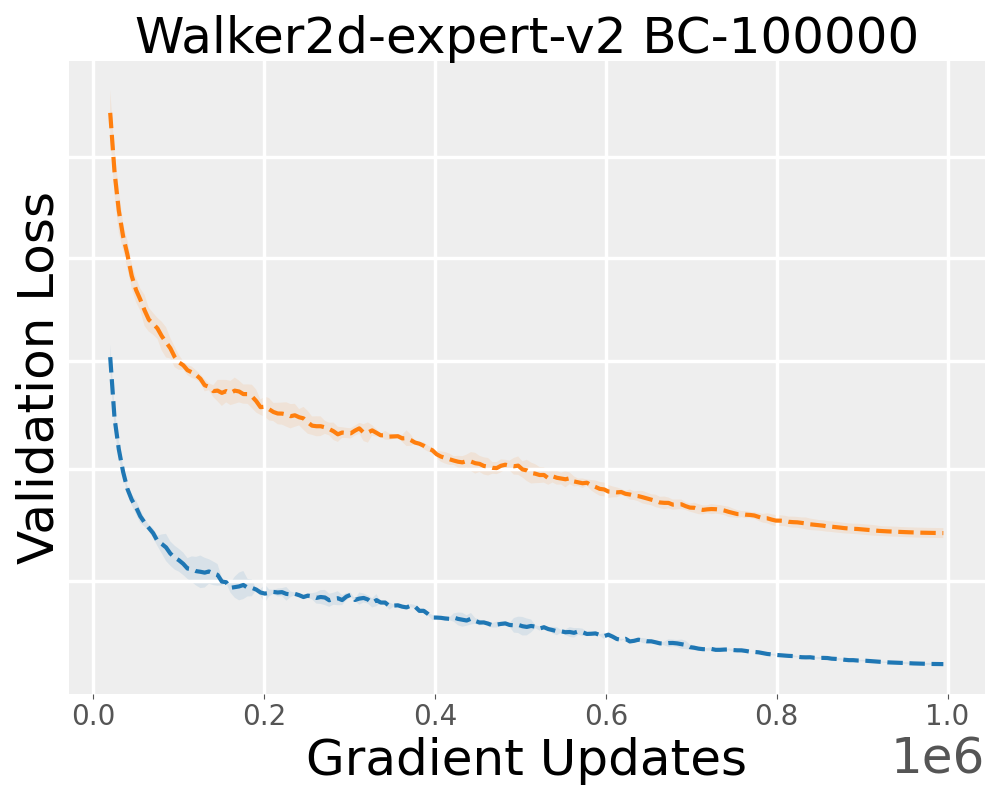}
\caption{\texttt{Walker2d-v2} trained with Expert dataset}
\end{subfigure}

\vspace{-6pt}
\caption{Compare the MSE loss of behaviour-action prediction on the \textbf{validation dataset} with varying training sample size between BC-baseline (\textcolor{cyan}{blue}) and sparse regularized BC (\textcolor{orange}{orange}) over 1 million training steps.}
\label{fig:validation_bc_loss}
\end{figure*}

\paragraph{Further Evidence of Reduced Overfitting:} 
In this controlled experiment, we train only the policy network with behaviour cloning (BC) loss. In Figure \ref{fig:training_bc_loss} and Figure \ref{fig:validation_bc_loss}, we present evidence of overfitting across various scenarios, including different range training samples ($5k$, $10k$, $50k$ and $100k$), datasets (medium and expert D4RL \citep{D4RL}), and multiple environments (\texttt{HalfCheetah-v2}, \texttt{Hopper-v2}, \texttt{Walker2d-v2}). The declining curves of the BC training loss (\textcolor{cyan}{blue}) in Figure \ref{fig:training_bc_loss} suggest an improvement in performance as the number of training samples increases. However, the validation curve (\textcolor{orange}{orange}) in Figure \ref{fig:validation_bc_loss} exhibits an upward trend, particularly for $5k$ and $10k$ training samples, indicating a loss of performance. Conversely, when considering BC-Sparse-Reg, the training loss does not decrease as dramatically as BC, but it demonstrates greater resilience in terms of validation loss. 


\begin{figure*}[hbt!]
\vspace{-10pt}
\centering

\begin{subfigure}{\textwidth}
\centering
\includegraphics[width=0.22\textwidth]{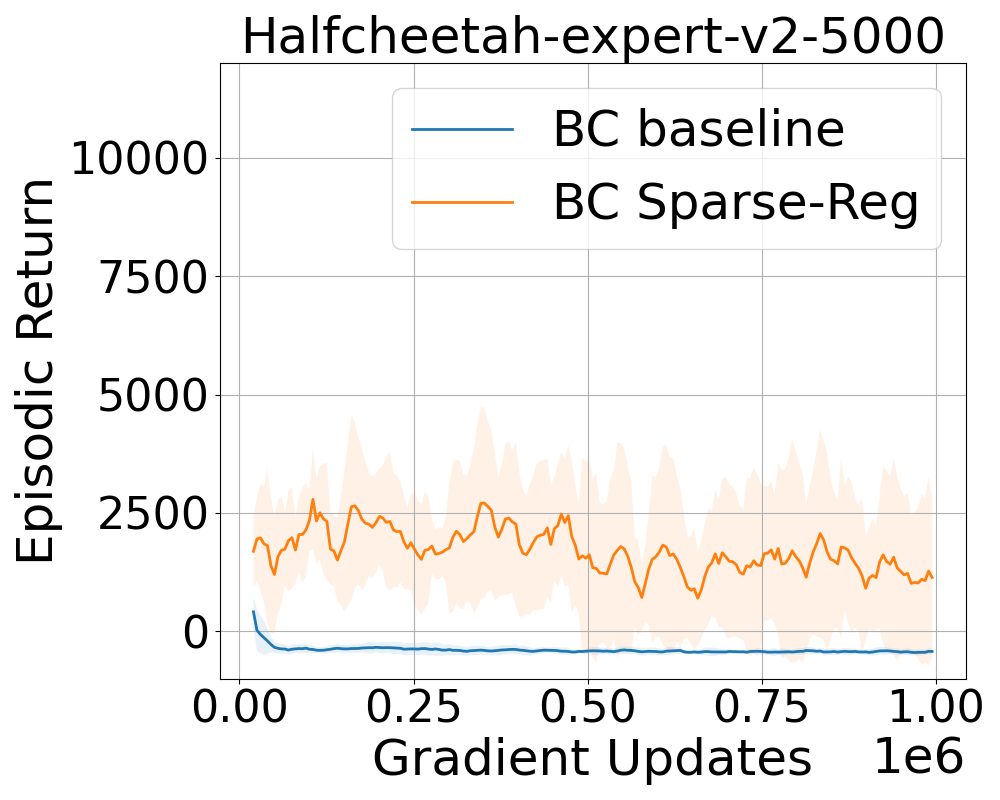}
\includegraphics[width=0.22\textwidth]{figures/BC/BC_compare_baseline_learning_curve/halfcheetah-expert-v2_10000_return.png}
\includegraphics[width=0.22\textwidth]{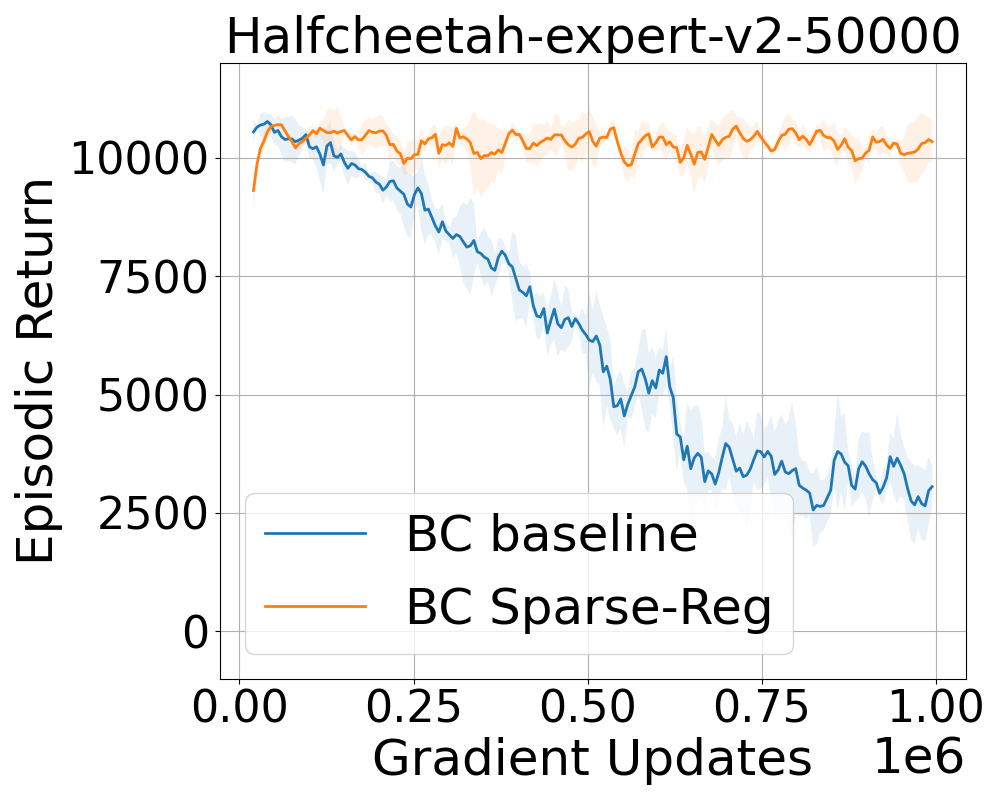}
\includegraphics[width=0.22\textwidth]{figures/BC/BC_compare_baseline_learning_curve/halfcheetah-expert-v2_100000_return.png}
\caption{\texttt{HalfCheetah-v2} trained with Expert dataset}
\end{subfigure}

\vspace{0.5em}

\begin{subfigure}{\textwidth}
\centering
\includegraphics[width=0.22\textwidth]{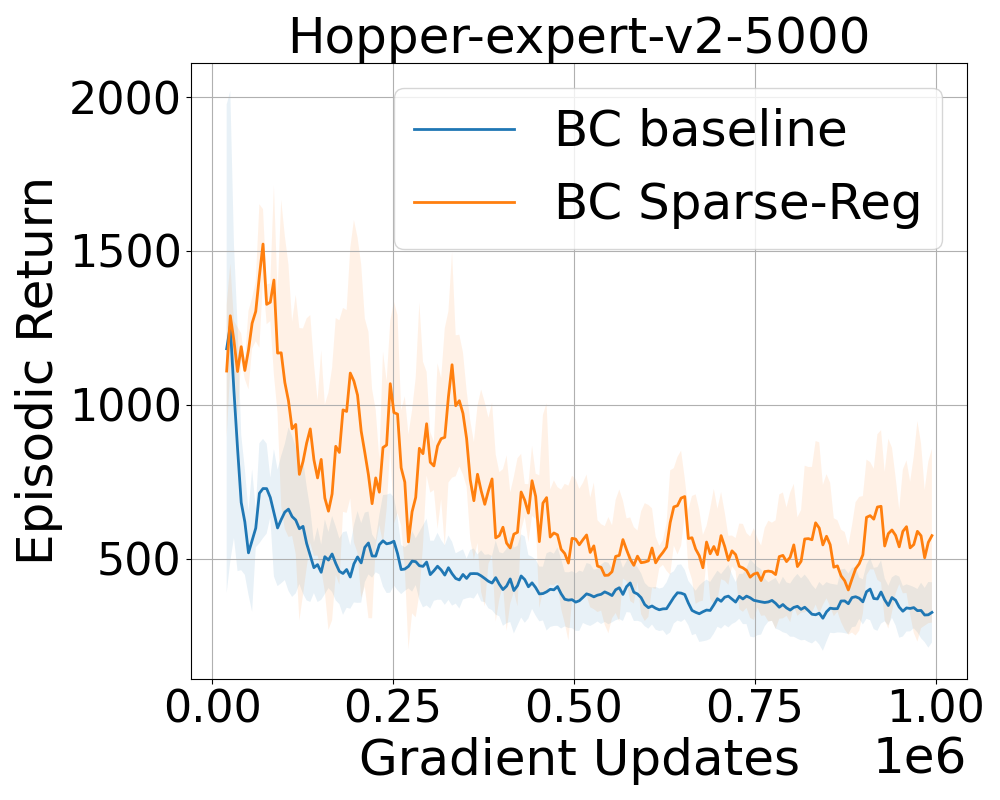}
\includegraphics[width=0.22\textwidth]{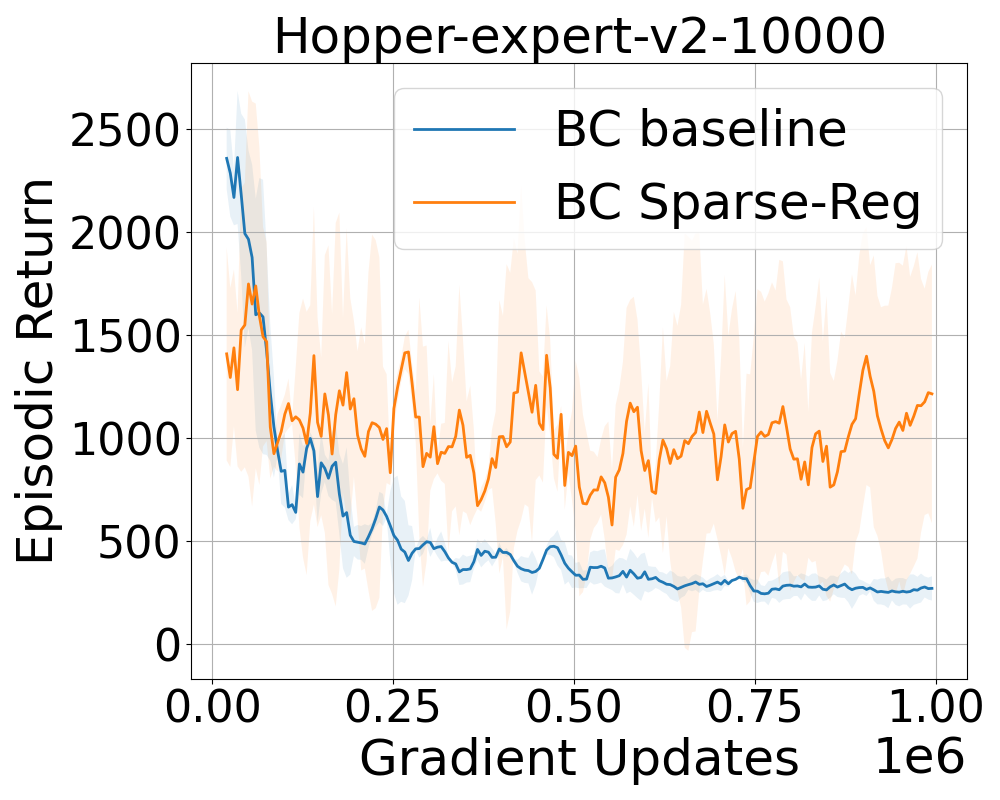}
\includegraphics[width=0.22\textwidth]{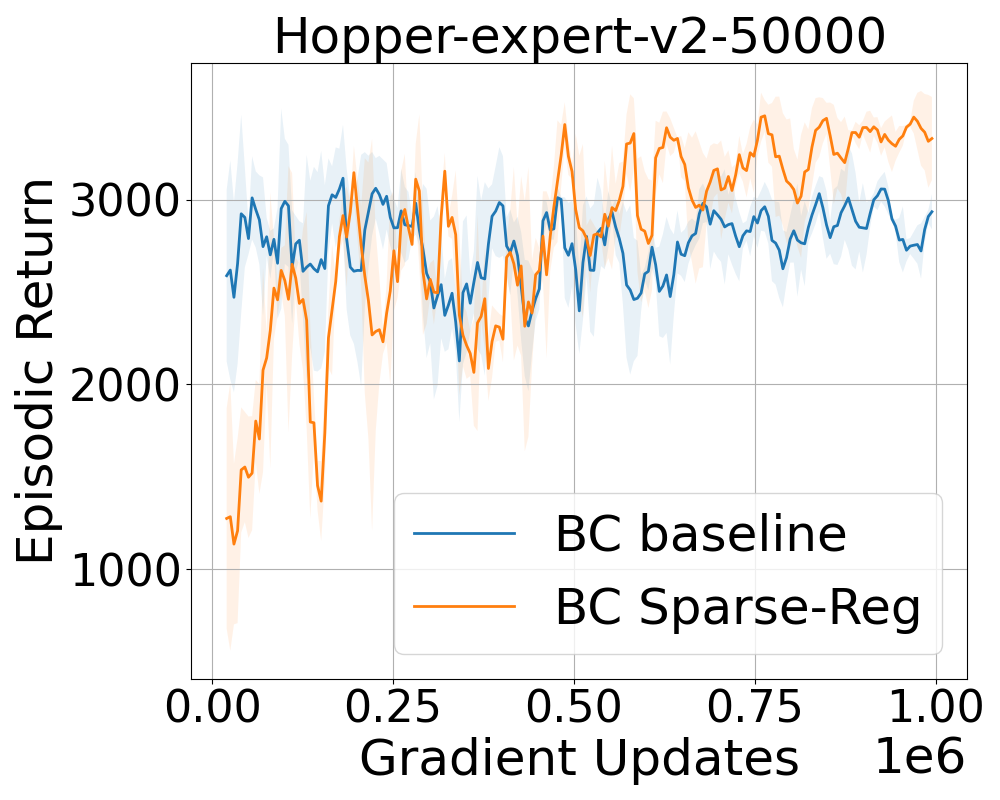}
\includegraphics[width=0.22\textwidth]{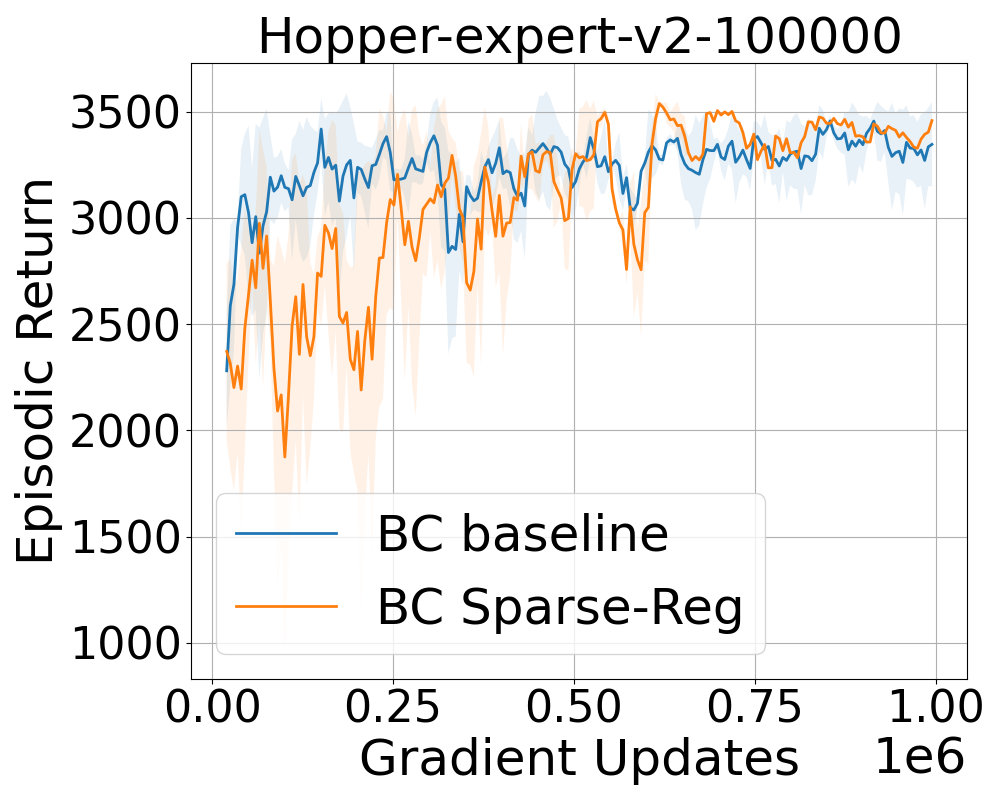}
\caption{\texttt{Hopper-v2} trained with Expert dataset}
\end{subfigure}

\vspace{0.5em}

\begin{subfigure}{\textwidth}
\centering
\includegraphics[width=0.22\textwidth]{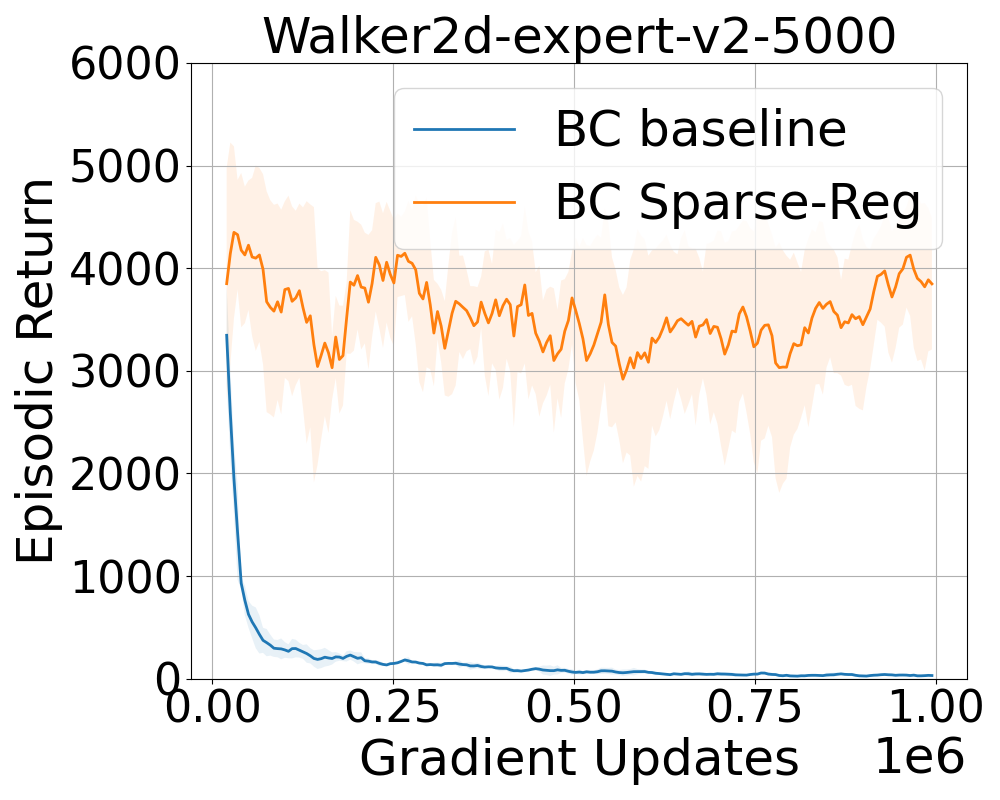}
\includegraphics[width=0.22\textwidth]{figures/BC/BC_compare_baseline_learning_curve/walker2d-expert-v2_10000_return.png}
\includegraphics[width=0.22\textwidth]{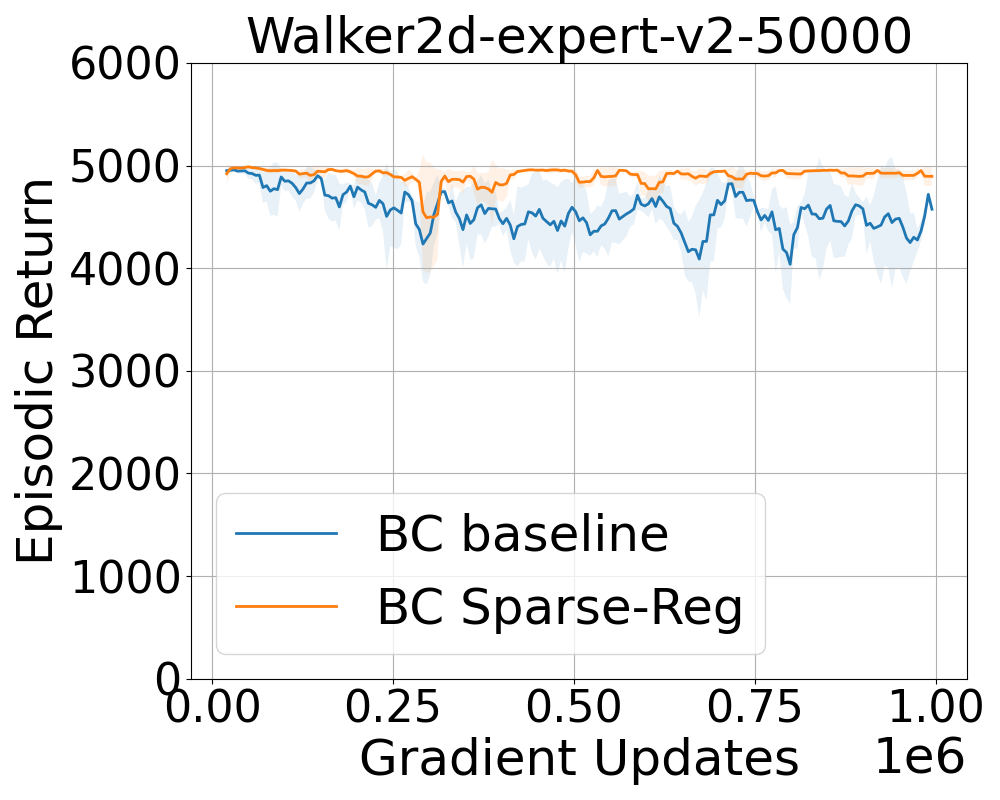}
\includegraphics[width=0.22\textwidth]{figures/BC/BC_compare_baseline_learning_curve/walker2d-expert-v2_100000_return.png}
\caption{\texttt{Walker2d-v2} trained with Expert dataset}
\end{subfigure}

\caption{Learning curve of BC-baseline (\textcolor{cyan}{blue}) and BC-sparse-reg (\textcolor{orange}{orange}) for ($a$) \texttt{HalfCheetah-v2}, ($b$) \texttt{Hopper}, ($c$) \texttt{Walker2d-v2} with varying training sample over 1 million time steps. Performance mean and standard deviation of the episodic returns are estimated every $5k$ steps over 10 evaluations across 5 seeds.}
\label{fig:appendix:BC_baseline_learning_curve}
\end{figure*}

\paragraph{Additional Experimental Results for Behavior Cloning:} Learning curve of BC varying training dataset size on \texttt{HalfCheetah}, \texttt{Hopper}, \texttt{Walker-2d} on Expert-dataset presented in Figure \ref{fig:appendix:BC_baseline_learning_curve}.

%% file: files/highlight_medium.tex
\begin{table*}[!]
\small
\caption{Performance improvement (without sparsity $\rightarrow$ with sparsity) of offline RL algorithms in D4RL benchmark ({\bf Medium}) varying dataset size. $\pm$ captures the standard deviation across 5 seeds. \sethlcolor{mine}\hl{Performance improvements} are highlighted blue and \sethlcolor{red!25}\hl{performance losses} are highlighted red. No performance loss is significant.}
\label{tab:offlineRL_baseline_medium}
\centering
\begin{tabularx}{0.99\textwidth}{cl*3{>{\raggedright\arraybackslash}X}}
\toprule
& & HalfCheetah & Hopper & Walker2d \\
\midrule

\multirow{6}{*}{\rotatebox[origin=c]{90}{Dataset Size: 5k}} 
& \multirow{2}{*}{TD3+BC}
  & $516.5 \pm 533.62$
  & $1388.11 \pm 113.95$
  & $764.45 \pm 130.52$ \\
& 
  & $\rightarrow$ \colorbox{mine}{$4047.8 \pm 224.79$}
  & $\rightarrow$ \colorbox{mine}{$1456.69 \pm 217.9$}
  & $\rightarrow$ \colorbox{mine}{$1640.83 \pm 587.8$} \\
\cmidrule{2-5}
& \multirow{2}{*}{AWAC}
  & $3217.39 \pm 260.64$
  & $329.76 \pm 106.56$
  & $30.91 \pm 35.55$ \\
& 
  & $\rightarrow$ \colorbox{mine}{$4576.05 \pm 233.81$}
  & $\rightarrow$ \colorbox{mine}{$1412.18 \pm 171.63$}
  & $\rightarrow$ \colorbox{mine}{$838.58 \pm 789.43$} \\
\cmidrule{2-5}
& \multirow{2}{*}{IQL}
  & $3699.08 \pm 326.82$
  & $782.36 \pm 108.01$
  & $280.62 \pm 12.26$ \\
&
  & $\rightarrow$ \colorbox{mine}{$4473.01 \pm 219.3$}
  & $\rightarrow$ \colorbox{mine}{$1328.21 \pm 77.15$}
  & $\rightarrow$ \colorbox{mine}{$1455.99 \pm 496.05$} \\
\midrule\midrule

\multirow{6}{*}{\rotatebox[origin=c]{90}{Dataset Size: 10k}} 
& \multirow{2}{*}{TD3+BC}
  & $1146.07 \pm 1076.7$
  & $1385.81 \pm 496.51$
  & $665.41 \pm 714.28$ \\
& 
  & $\rightarrow$ \colorbox{mine}{$3293.21 \pm 1970.12$}
  & $\rightarrow$ \colorbox{mine}{$1834.57 \pm 41.9$}
  & $\rightarrow$ \colorbox{mine}{$1050.78 \pm 820.76$} \\
\cmidrule{2-5}
& \multirow{2}{*}{AWAC}
  & $948.1 \pm 124.64$
  & $438.79 \pm 102.88$
  & $48.3 \pm 30.18$ \\
&
  & $\rightarrow$ \colorbox{mine}{$4832.04 \pm 84.98$}
  & $\rightarrow$ \colorbox{mine}{$1488.19 \pm 45.94$}
  & $\rightarrow$ \colorbox{mine}{$2367.16 \pm 743.22$} \\
\cmidrule{2-5}
& \multirow{2}{*}{IQL}
  & $2809.54 \pm 619.73$
  & $709.67 \pm 107.7$
  & $227.84 \pm 90.17$ \\
&
  & $\rightarrow$ \colorbox{mine}{$4653.05 \pm 236.0$}
  & $\rightarrow$ \colorbox{mine}{$1567.01 \pm 96.15$}
  & $\rightarrow$ \colorbox{mine}{$2123.09 \pm 216.2$} \\
\midrule\midrule

\multirow{6}{*}{\rotatebox[origin=c]{90}{Dataset Size: 50k}} 
& \multirow{2}{*}{TD3+BC}
  & $5287.29 \pm 116.86$
  & $1933.9 \pm 94.0$
  & $1983.34 \pm 724.94$ \\
&
  & $\rightarrow$ \colorbox{mine}{$5487.84 \pm 26.28$}
  & $\rightarrow$ \colorbox{red!25}{$1867.36 \pm 141.79$}
  & $\rightarrow$ \colorbox{mine}{$2872.2 \pm 457.49$} \\
\cmidrule{2-5}
& \multirow{2}{*}{AWAC}
  & $4634.02 \pm 114.0$
  & $1345.98 \pm 60.45$
  & $2213.39 \pm 285.77$ \\
&
  & $\rightarrow$ \colorbox{mine}{$4863.47 \pm 25.76$}
  & $\rightarrow$ \colorbox{mine}{$1615.1 \pm 27.7$}
  & $\rightarrow$ \colorbox{mine}{$2797.53 \pm 292.53$} \\
\cmidrule{2-5}
& \multirow{2}{*}{IQL}
  & $4648.26 \pm 59.31$
  & $1373.92 \pm 96.77$
  & $2215.77 \pm 192.56$ \\
&
  & $\rightarrow$ \colorbox{mine}{$4912.88 \pm 24.73$}
  & $\rightarrow$ \colorbox{mine}{$1593.18 \pm 83.48$}
  & $\rightarrow$ \colorbox{mine}{$2671.42 \pm 502.49$} \\
\midrule\midrule

\multirow{6}{*}{\rotatebox[origin=c]{90}{Dataset Size: 100k}} 
& \multirow{2}{*}{TD3+BC}
  & $3826.2 \pm 2922.24$
  & $1808.11 \pm 63.17$
  & $2416.76 \pm 1517.24$ \\
&
  & $\rightarrow$ \colorbox{mine}{$5538.44 \pm 52.42$}
  & $\rightarrow$ \colorbox{red!25}{$1796.81 \pm 82.88$}
  & $\rightarrow$ \colorbox{mine}{$3443.61 \pm 147.86$} \\
\cmidrule{2-5}
& \multirow{2}{*}{AWAC}
  & $4871.56 \pm 33.19$
  & $1505.84 \pm 73.18$
  & $1887.47 \pm 208.46$ \\
&
  & $\rightarrow$ \colorbox{mine}{$4953.73 \pm 22.84$}
  & $\rightarrow$ \colorbox{mine}{$1666.71 \pm 18.3$}
  & $\rightarrow$ \colorbox{mine}{$3344.39 \pm 27.58$} \\
\cmidrule{2-5}
& \multirow{2}{*}{IQL}
  & $4895.11 \pm 24.62$
  & $1533.8 \pm 24.36$
  & $2303.27 \pm 644.69$ \\
&
  & $\rightarrow$ \colorbox{mine}{$4924.37 \pm 54.48$}
  & $\rightarrow$ \colorbox{mine}{$1638.28 \pm 38.76$}
  & $\rightarrow$ \colorbox{mine}{$2829.0 \pm 660.87$} \\
\bottomrule
\end{tabularx}
\end{table*}

%% file: files/09_supplimentary.tex
\section{Additional Implementation Details}

\subsection{Sample Python Code}
Sparsity can be induced to any offline RL algorithm in a few lines of code. Here is a Python example code snippet:

\begin{figure}[htbp]
\centering
\scalebox{1.0}{
   \includegraphics[width=\linewidth]{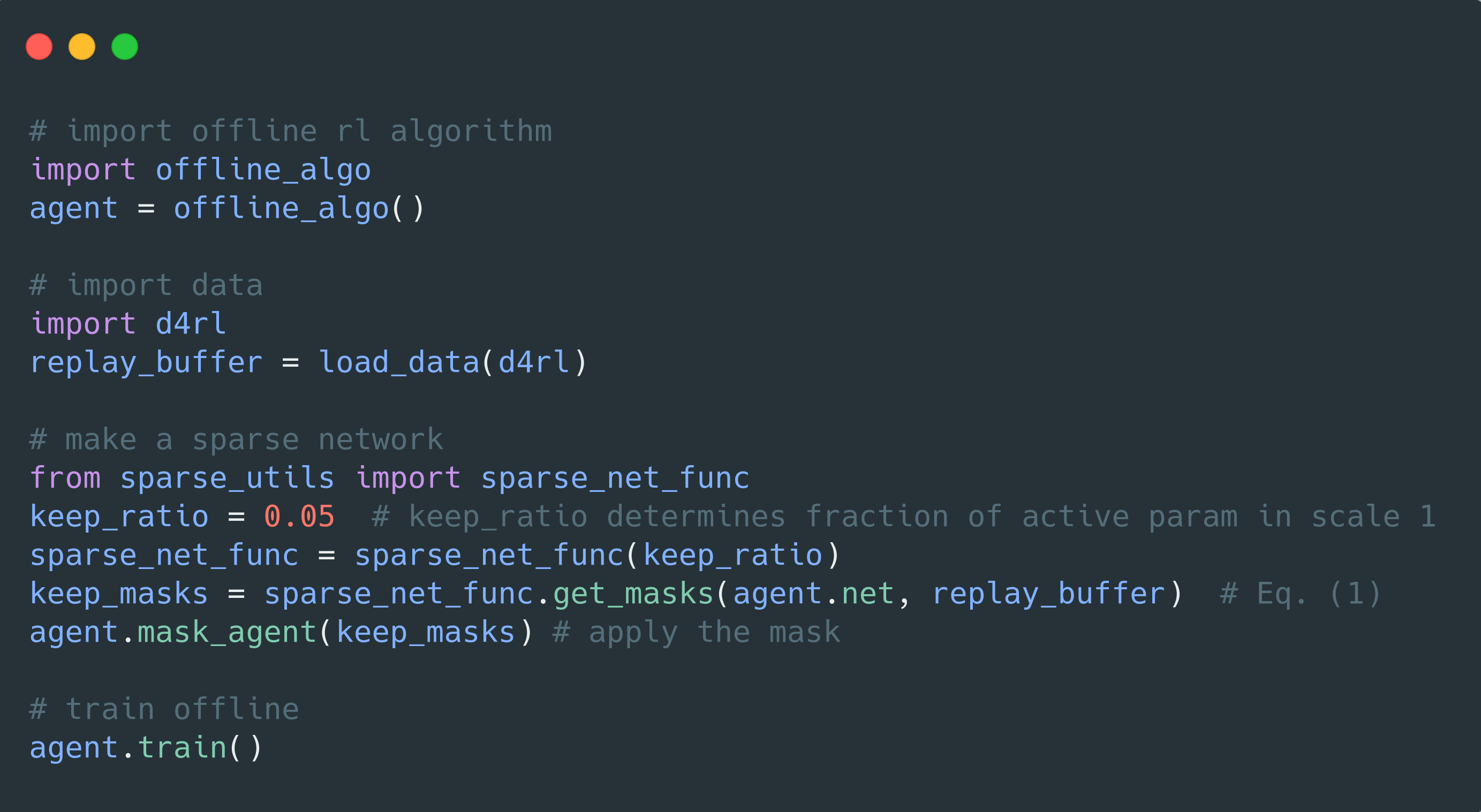}
}
  \vspace{-6pt}
  \caption{Example code of applying sparsity.}
\label{fig:code_snippet}
\end{figure}





\subsection{Libraries}

We run our algorithm in PyTorch-1.9.0 \cite{pytorch_lib} and use following libraries: TD3-BC \cite{TD3_BC}, AWAC \cite{CORL}, Implicit Q-learning (IQL) \cite{iql_pytorch} implementation. 

\subsection{Alternate Regularizer Details}
We utilize PyTorch \cite{pytorch_lib} implementation of different regularizers in our experiments. We apply \textbf{Dropout} \cite{dropout} and \textbf{Layer Norm} \cite{layer_norm} for each layer except the last layer for our experiment. We use a dropout rate of 0.1. Following the best practice from \cite{spectral_norm}, we apply \textbf{Spectral Norm} \cite{spectral_norm} at the penultimate layer of a neural network. To incorporate \textbf{Weight Decay} we apply AdamW \cite{AdamW} with a PyTorch default weight decay coefficient of 0.01.

\subsection{Hyper-parameters}
We use the default hyper-parameters used in the Offline algorithms. In Table \ref{table:networks_hyper_params} we present the network hyper-parameters of different algorithms that are used in this work.

\FloatBarrier
\begin{table}[!t]
\caption{Hyperparameter of the network architecture used to train and evaluate offline RL experiments.}
  \label{table:networks_hyper_params}
  \centering
  \scalebox{0.8}{
  \begin{tabular}{llllll}
    \toprule
     & Hyper-parameter & BC & TD3-BC & AWAC & IQL \\
    \midrule
hyper-parameter   &  Optimizer              &  Adam   & Adam & Adam & Adam \\
                  &  Critic learning rate   &  1e-3   & 1e-3  & 1e-3 & 1e-3  \\
                  &  Actor learning rate    &  1e-3   & 1e-3  & 1e-3 & 1e-3\\
                  & Mini-batch size         &  256    & 256     & 256  & 256\\
                  & Discount factor         &  0.99   & 0.99    & 0.99 & 0.99\\
                  & Target update rate      &  5e-3    & 5e-3     & 5e-3 & 5e-3 \\
                  & Policy update frequency &  2       & 2 & 2 & 2\\
    \midrule 
Architecture     &  Critic hidden dim       &  --  & [256, 256] & [256, 256] & [256, 256]\\
                 &  Critic activation function  & -- & ReLU & ReLU & ReLU \\
                 &  Actor hidden dim        &  [256,256] & [256,256] &  [256, 256] & [256, 256]\\
                 &  Actor activation function  & ReLU   & ReLu & ReLU & ReLU\\
                 &  Value hidden dim          &  -- & --&  -- & [256, 256]\\
                 &  Value activation function dim      & -- &   --  & -- & ReLU\\
    
    \bottomrule
  \end{tabular}
  }
\end{table}